\documentclass[journal]{IEEEtran}
\usepackage{epsfig}
\usepackage{graphicx}
\usepackage{amsmath}
\usepackage{amssymb}
\usepackage{epstopdf}
\usepackage{algorithm}
\usepackage{algpseudocode}
\usepackage{stackengine}

\ifCLASSINFOpdf
\else
\fi

\begin{document}	%
	\title{SymmSLIC: Symmetry Aware Superpixel Segmentation and its Applications}
	%
	%
	%
	
\author{Rajendra~Nagar
		and~Shanmuganathan~Raman\thanks{Rajendra Nagar and Shanmuganathan Raman are with the Electrical Engineering, Indian Institute of Technology, Gandhinagar,
			Gujarat, 382355, India e-mail:\{\texttt{rajendra.nagar, shanmuga}\}@iitgn.ac.in. \textit{A part of this paper is under consideration at Pattern Recognition Letters.}}
	}

	\maketitle
    
    \begin{abstract}
    	Over-segmentation of an image into superpixels has become a useful tool for solving various problems in image processing and computer vision. Reflection symmetry is quite prevalent in both natural and man-made objects and is an essential cue in understanding and grouping the objects in natural scenes. Existing algorithms for estimating superpixels do not preserve the reflection symmetry of an object which leads to different sizes and shapes of superpixels across the symmetry axis. In this work, we propose an algorithm to over-segment an image through the propagation of reflection symmetry evident at the pixel level to superpixel boundaries. In order to achieve this goal, we first find the reflection symmetry in the image and represent it by a set of pairs of pixels which are mirror reflections of each other. We partition the image into superpixels while preserving this reflection symmetry through an iterative algorithm. We compare the proposed method with state-of-the-art superpixel generation methods and show the effectiveness in preserving the size and shape of superpixel boundaries across the reflection symmetry axes. We also present two applications, symmetry axes detection and  unsupervised symmetric object segmentation, to illustrate the effectiveness of the proposed approach.
    \end{abstract}
    

    \section{Introduction}
    \label{sec:intro}
    \textbf{Superpixels.} A superpixel is a collection of spatially proximal and visually similar pixels \cite{ren2003learning}. The similarity could be defined in terms of color, texture, etc. The superpixels are known to preserve the local image features such as object boundaries, their regular shape and size, simple connectivity, and reduce the cost of computation of many computer vision problems. This is due to the fact that superpixel over-segmentation effectively reduces the number of units to be processed in an image. The superpixel segmentation has been used in applications such as segmentation \cite{ren2003learning}, image parsing \cite{tighe2010superparsing}, tracking \cite{wang2011image}, and 3D reconstruction \cite{hoiem2005automatic}.
    \\
    \textbf{Symmetry.} The symmetry present in real-world objects is proven to play a major role in object detection and object recognition processes in humans and animals \cite{tyler2003human}. Therefore, detecting the symmetry evident in the objects has become an important area of research. The major types of symmetry are reflection symmetry, rotation symmetry, and translation symmetry. The most commonly occurring symmetry in nature is the reflection symmetry. We mainly focus on preserving the reflection symmetry present in the image. The reflection symmetry present in natural images has been used to solve many problems in computer vision such as object detection \cite{sun2012reflection}, image matching \cite{hauagge2012image}, facial images analysis \cite{mitra2004local}, real-time attention for robotic vision \cite{sela1997real}, tumor segmentation in medical images \cite{mancas2005fast}, 3D reconstruction  \cite{koser2011dense,sinha2012detecting}, shape manipulation \cite{ming2013symmetry}, model compression \cite{ming2013symmetry}, and symmetrization \cite{ming2013symmetry}. The common way to represent reflection symmetry is through a set of pairs of mirror symmetric  pixels and the axis of symmetry.
    \begin{figure*}[htbp]
    	\centering
    	
    				\epsfig{figure= 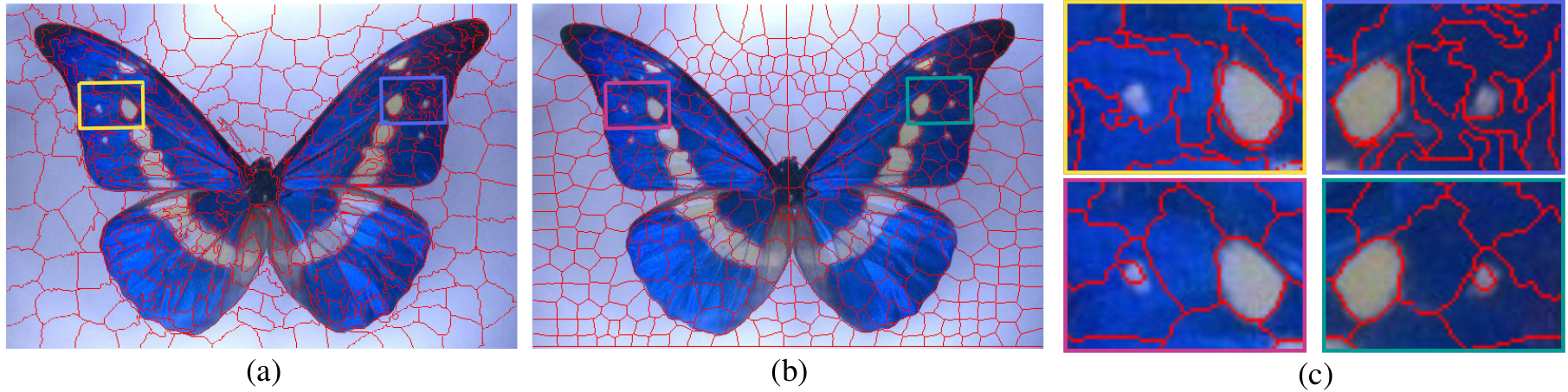,width=1\linewidth}
    		
    				\caption{(a) The results for the Manifold SLIC \cite{liu2016manifold}, (b) results for the proposed algorithm SymmSLIC, and (c) zoomed mirror symmetric windows for both the methods (top: MSLIC, bottom: ours). We observe that the symmetry at the superpixel level is preserved in ours.}
    				\label{fig:RI_1}
    			\end{figure*}
    			\\
    			\textbf{Motivation.} The perceptual grouping of local object features is a major cue in understanding objects in the human visual system \cite{wertheimer1938laws}. The symmetry present in real-world objects is proven to play a major role in object detection and object recognition processes in humans as well as animals \cite{tyler2003human}. Therefore, the symmetry present in objects should be preserved even after perceptual grouping in order to perceive the objects efficiently from the perceptually similar groups. The existing superpixel algorithms do not attempt to preserve the symmetry present in the image. The main motivation behind preserving symmetry at superpixel level is that the time complexity of algorithms using symmetry such as \cite{hauagge2012image} can be reduced significantly by working at superpixels level. However, without preserving the symmetry at superpixels, their performance might get degraded. There have been attempts in preserving structure \cite{wang2013structure,liu2016manifold}. However, no emphasis has been made on preserving symmetry. In this work, we propose an algorithm to partition an image into superpixels while preserving the reflection symmetry. At the superpixel level, we represent the symmetry as a set of pairs of superpixels which are mirror reflections of each other.  We improve and extend the SLIC algorithm to achieve this task \cite{achanta2012slic}. In Figure \ref{fig:RI_1}, we show an example output generated by the proposed approach along with another recent superpixel segmentation method Manifold-SLIC for illustration \cite{liu2016manifold}. The main contributions of this work are the following.
    			\begin{enumerate}
    				\item We propose an algorithm, termed \textit{SymmSLIC}, in order to partition an image into superpixels such that the reflection symmetry present at the pixel level is preserved at the superpixel level.
    				\item We also propose a novel algorithm to detect pairs of pixels which are mirror reflections of each other. 
    				\item We introduce an application of SymmSLIC called unsupervised symmetric object segmentation and exploit detected pairs of mirror symmetric pixels to detect symmetry axes  present in the input image.
    			\end{enumerate}
    			This article is an extended and revised version of the conference paper \cite{Nagar_2017_ICCV_Workshops}.
    			
    			We organize the remainder of the paper as follows. In Section \ref{sec:RW}, we present literature review of superpixel segmentation and symmetry axis detection. In Section \ref{subsec:RA}, we discuss the proposed method for the detection of mirror symmetry point pairs. In Section \ref{subsec:slic}, we develop the SymmSLIC algorithm. In Section \ref{sec:RD}, we present the results for reflection symmetry aware superpixel segmentation. In Section \ref{sec:5}, we provide two applications:  unsupervised symmetric object segmentation and symmetry axes detection. In Section \ref{sec:c}, we conclude the paper with discussion on limitations and future directions.
    			\section{Related Works}
    			\label{sec:RW}
    			Superpixel segmentation and symmetry detection problems have been studied thoroughly and are active research problems in image processing, computer vision, and computer graphics. To the best of our knowledge, there have not been any previous attempt on the problem of symmetry preserving superpixel segmentation. We discuss the state-of-the-art methods for superpixel segmentation and symmetry detection methods. 
    			\\
    			\textbf{Superpixels.} There are two major categories of algorithms for superpixel segmentation - graph based and clustering based. Following are the major graph based approaches. Shi and Malik proposed normalized cut algorithm to over-segment an image \cite{shi2000normalized}. Felzenszwalb proposed a graph-based image segmentation approach \cite{felzenszwalb2004efficient}. Li and Chen used linear spectral clustering approach \cite{li2015superpixel}.  \cite{veksler2010superpixels} and \cite{van2015seeds} used optimization techniques for superpixel segmentation. Zhang \emph{et al.} proposed a boolean optimization framework for superpixel segmentation \cite{zhang2011superpixels}. Moore \emph{et al.}  posed the problem of superpixel detection as lattice detection \cite{moore2008superpixel}. Duan and Lafarge used shape anchoring techniques on the set of detect line segments in order to partition the image into convex superpixels \cite{duan2015image}. The key idea in the clustering based approaches is to first initialize some cluster centers and then refine these cluster centers using various techniques. Achanta \emph{et al.} proposed a k-means clustering based approach called simple linear iterative clustering (SLIC) \cite{achanta2012slic}. They initialize cluster centers at the centers of equally spaced squares. They perform clustering by assigning each pixel to the nearest center based on the color and location similarity. Levinshtein \emph{et al.} proposed a geometric flow based approach \cite{levinshtein2009turbopixels}. Wang \emph{et al.} proposed a content sensitive superpixel segmentation approach where the distance between the cluster center and a pixel is the geodesic distance \cite{wang2013structure}. Liu \emph{et al.} proposed a fast algorithm to get structure sensitive superpixels, where authors perform the SLIC on a 2-dimensional manifold \cite{liu2016manifold}. Liu \emph{et al.} used the entropy rate for homogeneous and compact superpixels \cite{liu2011entropy}. 
    			\\
    			\textbf{Reflection Symmetry.} The problem of detecting reflection symmetry present in images have been thoroughly studied recently \cite{liu2010computational,ming2013symmetry}. The existing approaches for symmetry detection in images can be categorized in four categories - \textit{direct approach}  \cite{vasilier1984recognition,krahe1986detection}, \textit{voting based approaches} \cite{parui1983symmetry,yip2000hough,mitra2006partial,loy2006detecting,atadjanov2016reflection}, \textit{basis function based approaches} \cite{yodogawa1982symmetropy}, and \textit{moment based approaches}  \cite{marola1989detection,shen1999symmetry}. Loy and Eklundh mirrored the scale invariant feature transform (SIFT \cite{lowe2004distinctive}) descriptors in order to get the reflection invariant SIFT descriptors \cite{lowe2004distinctive}. In order to match two points they used SIFT descriptor for one point and mirrored-SIFT descriptor for the other point. Then, they detect the symmetry axis using Hough transform based line detection algorithm \cite{loy2006detecting}.  Kondra \emph{et al.} proposed a kernel based approach \cite{kondra2013multi}. Patraucean \emph{et al.} used affine invariant edge features and a \emph{contrarion} validation scheme \cite{patraucean2013detection}. Michaelsen \emph{et al.} used Gestalt algebra \cite{michaelsen2013recognition}. Atadjanov and Lee detected symmetry axes using appearance of structure features \cite{atadjanov2016reflection}. The works \cite{tsogkas2012learning,scognamillo2003feature,zielke1993intensity,sun1995symmetry,cho2009bilateral,bokeloh2009symmetry,ming2013symmetry,mo2011detecting,wang2015reflection,cornelius2006detecting,tuytelaars2003noncombinatorial,tang2014reflexive} present good methods for symmetry detection. There also have been works in detecting symmetry in 3D geometric models \cite{mitra2013symmetry}.

    			\section{Proposed Approach}
    			
    			\label{sec:PA}
    			\subsection{Approximate and Partial Reflection Symmetry Detection }
    			\label{subsec:RA}
    			Let $I:\mathcal{W}\times\mathcal{H}\rightarrow\mathbb{R}^3$ be a color image with width $w$ and height $h$, where $\mathcal{W}=\{1,2,\ldots,w\}$ and $\mathcal{H}=\{1,2,\ldots,h\}$. Most of the real images exhibit only the \textit{partial reflection symmetry}, which means that the mirror symmetric pixel exists for only a fraction of pixels. This is due to the fact that the images have square boundaries and the boundaries of the real object are not necessarily square and furthermore, there could be occlusions and missing parts. Since the real objects are not perfectly mirror symmetric, we attempt to detect the \textit{approximate reflection symmetry}. Our goal is to detect the partial and approximate reflection symmetry present in the input image $I$. We detect the pairs of pixels which are mirror reflections of each other. We represent the reflection symmetry present in the image by two subsets, $\mathcal{L}\subset\mathcal{W}\times\mathcal{H}$ and  $\mathcal{R}\subset\mathcal{W}\times\mathcal{H}$, satisfying the following property. 
    			For each pixel $\mathbf{x}_i\in\mathcal{L},\exists\; \mathbf{x}_{i^\prime}\in\mathcal{R}$ such that 
    			$$\mathbf{x}_{i^\prime}=\mathbf{R}_{ii^\prime}\mathbf{Q}\mathbf{R}_{ii^\prime}^\top (\mathbf{x}_i-\mathbf{t}_{ii^\prime})+\mathbf{t}_{ii^\prime},\text{ and }I(\mathbf{x}_i)=I(\mathbf{x}_{i^\prime}).$$
    			Here, the point $\mathbf{t}_{ii^\prime}=\frac{\mathbf{x}_i+\mathbf{x}_{i^\prime}}{2}$, the matrix $\mathbf{Q}=\begin{bmatrix}1 &0\\0&-1\end{bmatrix}$, and the matrix $\mathbf{R}_{ii^\prime}=\begin{bmatrix}\cos\theta_{ii^\prime}&-\sin\theta_{ii^\prime}\\\sin\theta_{ii^\prime}&\cos\theta_{ii^\prime}\end{bmatrix}$. The angle $\theta_{ii^\prime}$ is the slope of the symmetry axis which is a line perpendicular to the vector $\mathbf{x}_i-\mathbf{x}_{i^\prime}$ and passes through the mid-point $\mathbf{t}_{ii^\prime}$.  In order to determine the sets $\mathcal{L}$ and  $\mathcal{R}$, representing the reflection symmetry, we use normals of the edges present in the image. We first extract the edges from the image using \cite{arbelaez2011contour} and represent them as curves. Let $E:\mathcal{W}\times\mathcal{H}\rightarrow\{0,1\}$ be the image representing the edges present in the image $I$. Let $\mathcal{F}=\{\mathbf{x}:E(\mathbf{x})=1\}$ be the set of pixels lying on the edges. Now, for each pixel $\mathbf{x}_i \in \mathcal{F}$, we extract an edge of length $p$ pixels passing through the pixel $\mathbf{x}_i$ such that the pixel $\mathbf{x}_i$ lies at equal distance from the end points of the edge, and represent it by a curve. We represent it by the curve $c_{i}(\alpha):[0,1]\rightarrow\mathbb{R}^2$ such that $c_{i}(0.5)=\mathbf{x}_i$ with length $p$. We determine pairs of pixels which are mirror reflections of each other with a confidence score based on the following observation.
    			
    			Let $\mathbf{x}_i$, $\mathbf{x}_{i^\prime}\in\mathcal{F}$ be any two edge pixels. If the pixels $\mathbf{x}_i$ and $\mathbf{x}_{i^\prime}$ are mirror reflections of each other and the image $I$ is mirror symmetric in the proximity of the pixels $\mathbf{x}_i$ and $\mathbf{x}_{i^\prime}$, then the following equalities hold true. 
    			\begin{eqnarray}
    			c_{i^\prime}(\alpha)&= &\mathbf{R}_{ii^\prime}\mathbf{Q}\mathbf{R}_{ii^\prime}^\top (c_i(\alpha)-\mathbf{t}_{ii^\prime})+\mathbf{t}_{ii^\prime},\forall\alpha\in[0, 1]\\
    			c_i(\alpha)&=&\mathbf{R}_{ii^\prime}\mathbf{Q}\mathbf{R}_{ii^\prime}^\top (c_{i^\prime}(\alpha)-\mathbf{t}_{ii^\prime})+\mathbf{t}_{ii^\prime},\forall\alpha\in[0, 1].
    			\end{eqnarray}
    			
    			Let $\boldsymbol{\eta}_{i}(\alpha)$ be the normal to the curve $c_i(\alpha)$ at $\alpha$. It is trivial to show that if the curve $c_i$ and $c_{i^\prime}$ are mirror symmetric then $\boldsymbol{\eta}_{ii^\prime}(\alpha)=\boldsymbol{\eta}_{i^\prime}(\alpha)$ and $\boldsymbol{\eta}_{i^\prime i}(\alpha)=\boldsymbol{\eta}_{i}(\alpha)$. Here, $\boldsymbol{\eta}_{i^\prime i}(\alpha)$ is normal to the curve $\mathbf{R}_{ii^\prime}\mathbf{Q}\mathbf{R}_{ii^\prime}^\top (c_{i^\prime}(\alpha)-\mathbf{t}_{ii^\prime})+\mathbf{t}_{ii^\prime}$ at $\alpha$, and $\boldsymbol{\eta}_{ii^\prime}(\alpha)$ is normal to the curve $\mathbf{R}_{ii^\prime}\mathbf{Q}\mathbf{R}_{ii^\prime}^\top (c_i(\alpha)-\mathbf{t}_{ii^\prime})+\mathbf{t}_{ii^\prime}$ at $\alpha$.   
    			
    			Therefore, if the pixels $\mathbf{x}_i$ and $\mathbf{x}_{i^\prime}$ are mirror reflections of each other, then we have that  
    			\begin{equation} \int_{0}^{1}(\boldsymbol{\eta}_{i^\prime i}^\top(\alpha)\boldsymbol{\eta}_{i}(\alpha)+\boldsymbol{\eta}_{i i^\prime}^\top(\alpha)\boldsymbol{\eta}_{i}(\alpha))d\alpha=\int_{0}^{1}(1+1)d\alpha=2.
    			\label{eq3}
    			\end{equation}
    			
    			For the case of perfect symmetry, equation \ref{eq3} holds true. However, due to the presence of noise and illumination variations, this might not hold true in practice. Furthermore, there may be various outlier pairs. Our goal is to detect all the symmetry axes using the set of such pairs. First we cluster the pairs of mirror symmetric pixels.  And then, we find the symmetry axis in each cluster separately. Let $\{(\mathbf{x}_i,\mathbf{x}_i^\prime)\}_{i=1}^v$ be the detected pairs of mirror symmetric pixels. We construct an undirected graph $\mathcal{G}=(\mathcal{V},\mathcal{E})$, where each vertex $v_i$ in the vertex set $\mathcal{V}$ corresponds to the pair $(\mathbf{x}_i,\mathbf{x}_{i^\prime})$. We connect the vertices $v_i$ and $v_j$ by an unit weight edge if the symmetry axes $L_i$  and $L_j$, defined by the pairs $(\mathbf{x}_i,\mathbf{x}_{i^\prime})$ and $(\mathbf{x}_j,\mathbf{x}_{j^\prime})$ respectively, are similar. We define the similarity between the lines as follows. Let $c_{j^\prime i}(\alpha)=\mathbf{R}_{ii^\prime}\mathbf{Q}\mathbf{R}_{ii^\prime}^\top( c_{j}(\alpha)-\mathbf{t}_{ii^\prime})-\mathbf{t}_{ii^\prime}$ be the reflection of the curve through the symmetry axis $L_i$ and $\boldsymbol{\eta}_{j^\prime i}(\alpha)$ be its normal at $\alpha$.  Similarly, let $c_{i^\prime j}(\alpha)=\mathbf{R}_{jj^\prime}\mathbf{Q}\mathbf{R}_{jj^\prime}^\top( c_{i}(\alpha)-\mathbf{t}_{jj^\prime})-\mathbf{t}_{jj^\prime}$ be the reflection of the curve through the symmetry axis $L_j$ and $\boldsymbol{\eta}_{i^\prime j}(\alpha)$ be its normal at $\alpha$. If the symmetry axes $L_i$  and $L_j$ are similar then $\boldsymbol{\eta}_{i^\prime j}$ and $\boldsymbol{\eta}_{j^\prime i}$ will also be similar. Therefore, we create an edge between the vertices $v_i$ and $v_j$ if $\int_{0}^{1}(\boldsymbol{\eta}_{i^\prime j}^\top(\alpha)\boldsymbol{\eta}_{i^\prime}(\alpha)+\boldsymbol{\eta}_{j^\prime i}^\top(\alpha)\boldsymbol{\eta}_{j^\prime}(\alpha))d\alpha>2\tau$.
    			\begin{figure*}[htbp]
    				\centering
    				\stackunder{\epsfig{figure= 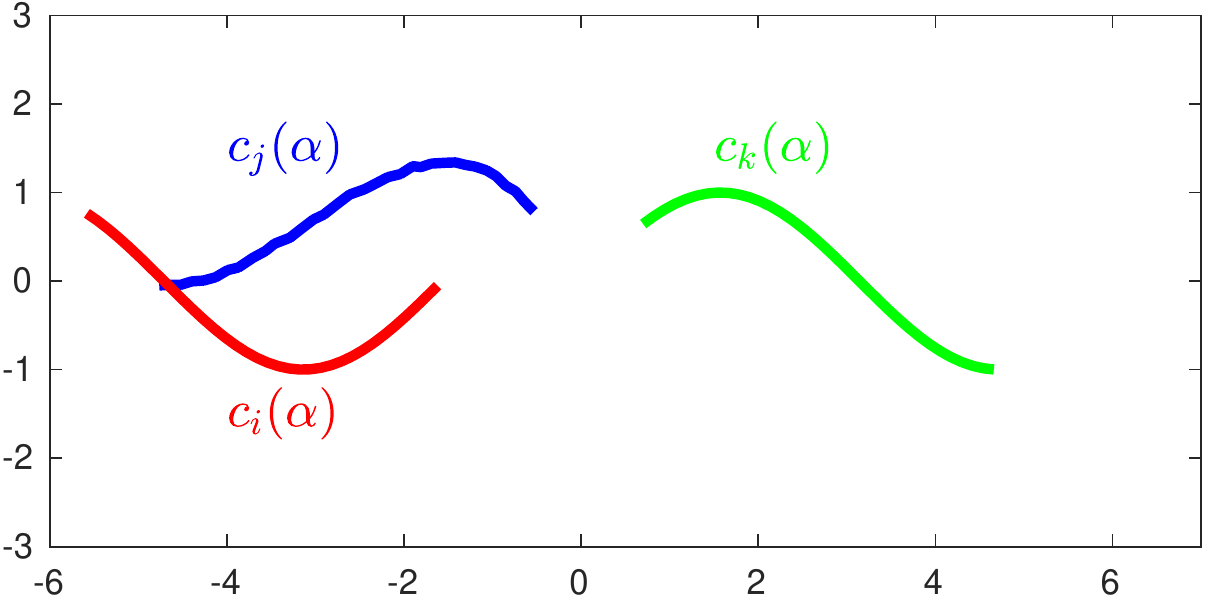,width=0.46\linewidth}}{(a)}
    				\stackunder{\epsfig{figure= 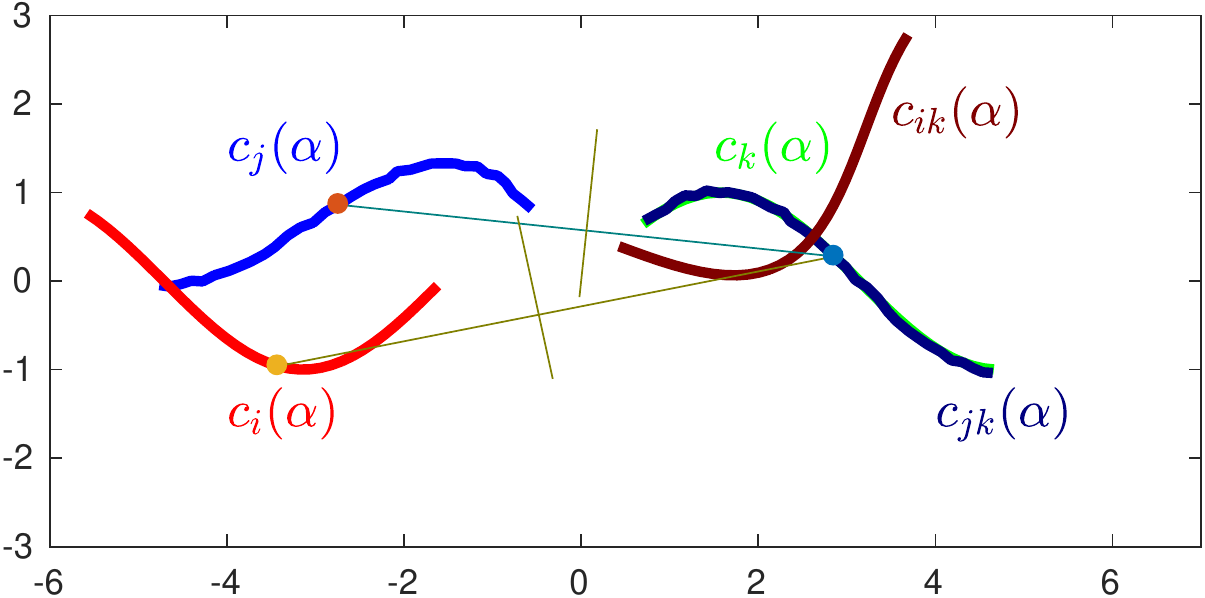,width=0.46\linewidth}}{(b)}
    				\stackunder{\epsfig{figure= 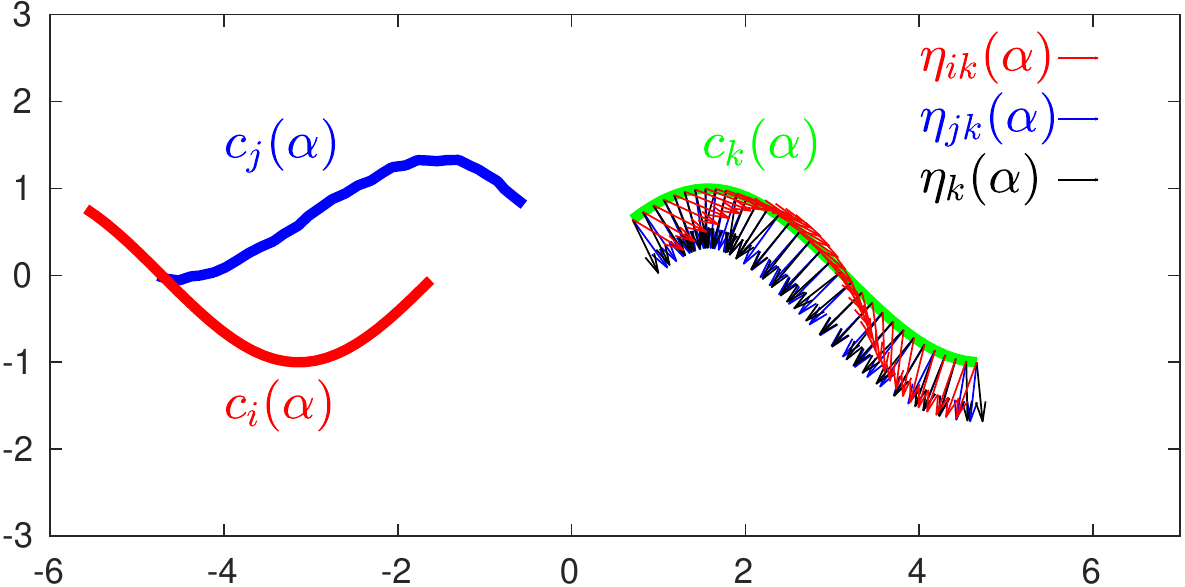,width=0.46\linewidth}}{(c)}
    				\stackunder{\epsfig{figure= 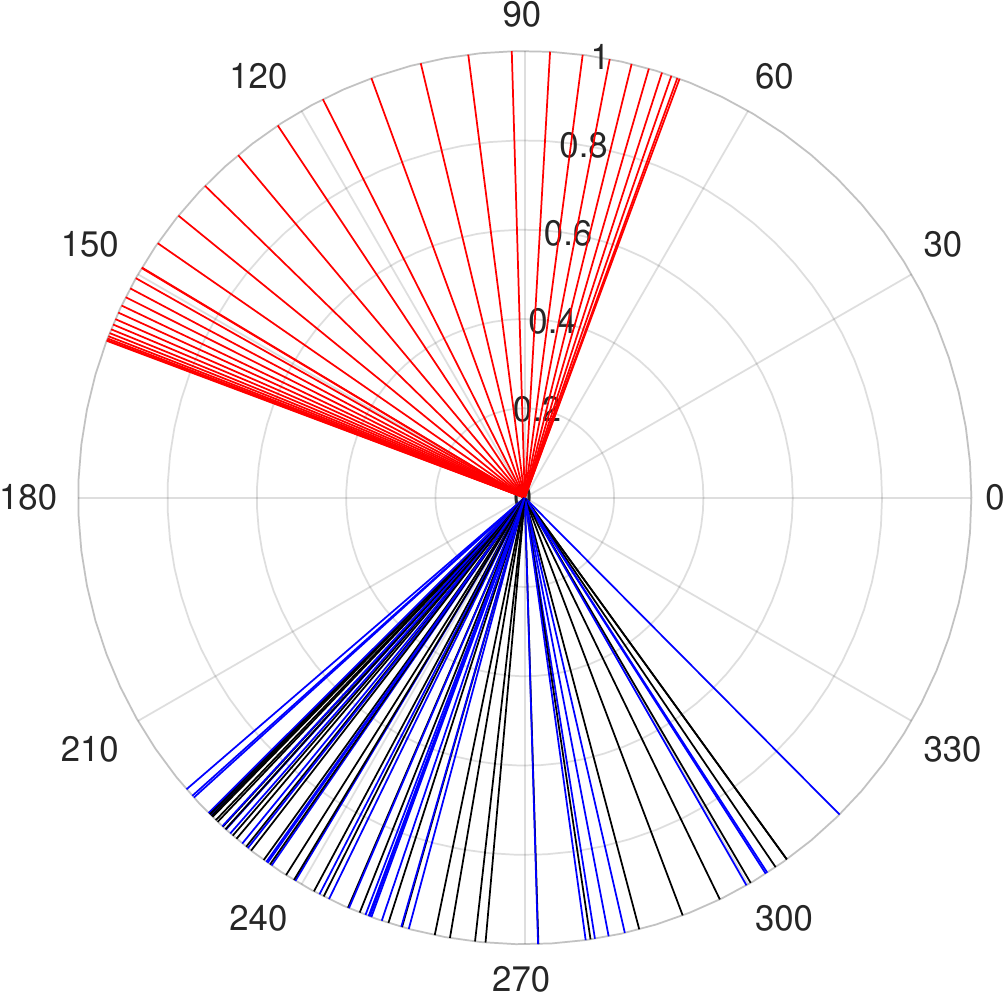,width=0.23\linewidth}}{(d)}
    				
    				\caption{Illustration of pairs of mirror symmetric pixels detection approach.
    					(a) Three curves $c_{i}(\alpha)$, $c_{j}(\alpha)$, and $c_{k}(\alpha)$. (b) Reflection of the curves $c_{i}(\alpha)$, $c_{j}(\alpha)$ about the symmetry axes $L_{ki}$ and $L_{ji}$ defined by the pairs $(c_{k},c_{i})$ and $(c_{k},c_j)$, respectively. (c)-(d) The normals $\boldsymbol{\eta}_{ki}$,  $\boldsymbol{\eta}_{kj}$, $\boldsymbol{\eta}_{k}$ to the curves $c_{ki}$, $c_{kj}$, and $c_{k}$, respectively,  shown on the curve $c_k$ for better comparison.}
    				\label{fig:mspp2}
    			\end{figure*}
    			
    			In Figure \ref{fig:mspp1} (a), we show a graphical illustration. We observe that each clique in the graph $\mathcal{G}$ corresponds to the set of pairs of mirror symmetric pixels belonging to a same symmetry axis \cite{kleinberg2006algorithm}. A clique in the graph $\mathcal{G}$ is a subset $\mathcal{C}$ of the vertex set $\mathcal{V}$  such that every vertex in $\mathcal{C}$ is connected by an edge. Therefore, our goal is to find all the dominant cliques in the graph $\mathcal{G}$. It is a well known result that a clique is equivalent to an independent set in the complement graph. The complement graph $\bar{\mathcal{G}}=(\bar{\mathcal{V}},\bar{\mathcal{E}})$ of a graph $\mathcal{{G}}$ is the graph such that $\bar{\mathcal{V}}=\mathcal{V}$ , $(u,v)\in\mathcal{E}\Rightarrow(u,v)\notin\bar{\mathcal{E}}$, and $(u,v)\notin\mathcal{E}\Rightarrow(u,v)\in\bar{\mathcal{E}}$.  An independent set in the graph $\bar{\mathcal{G}}$ is a subset $\mathcal{I}$ of the vertex set $\bar{\mathcal{V}}$  such that no two vertices in  $\mathcal{I}$ are adjacent. Furthermore, the independent set is complement of the vertex cover. A vertex cover of an undirected graph $\bar{\mathcal{G}}$ is a subset $\mathcal{V}_\text{c}$ of vertices of $\bar{\mathcal{V}}$ such that if $(v_i,v_j)$ is an edge in $\bar{\mathcal{G}}$, then either $v_i\in\bar{\mathcal{V}}$ or $v_j\in\bar{\mathcal{V}}$. In order to find the minimum vertex cover, we solve the following integer linear program (ILP).
    			\begin{eqnarray}
    			\nonumber &\text{min }\sum_{v\in\bar{\mathcal{V}}}x_v&\\
    			\nonumber \text{s.t.} &x_u+x_v\geq1 &\forall (u,v)\in\bar{\mathcal{E}}\\
    			&    x_v\in\{0,1\}&\forall v\in\bar{\mathcal{V}}
    			\end{eqnarray}     
    			Here, the binary variable $x_v$ is equal to 1, if the vertex $v$ is in the vertex cover $\mathcal{V}_\text{c}$ and 0, otherwise. The constraint $x_u+x_v\geq1$ ensures that at least one vertex of the edge $(u,v)\in\bar{\mathcal{E}}$ is included in the vertex cover. We rewrite the above ILP in the standard form as below
    			\begin{eqnarray}
    			\underset{\mathbf{x}}{\text{argmin }}\mathbf{1}^\top\mathbf{x} \text{ subject to } \mathbf{Ex}\geq 1,\;\;\;    \mathbf{x}\in\{0,1\}^{|\bar{\mathcal{V}}|}.&
    			\label{eq6}
    			\end{eqnarray}
    			Here, $\mathbf{1}$ is a vector of size $|\bar{\mathcal{V}}|$ with all elements equal to 1 and the matrix $\mathbf{E}\in\{0,1\}^{|\bar{\mathcal{E}}|\times |\bar{\mathcal{V}}|}$ is the edge incident matrix such that $\mathbf{E}(e,v)=1$ if the $e$-th edge is incident on the vertex $v$ and zero, if the $e$-th edge is not incident on the vertex $v$. The vertex cover problem is an NP-hard problem. Therefore, we use the best known approximation which is 2-approximation obtained by relaxing the integer linear program in equation (\ref{eq6}) to a linear program. In the relaxed program, each variable takes value in $[0,1]$, i.e., $\mathbf{x}\in[0,1]^{|\bar{\mathcal{V}}|}$. We obtain the final solution by an optimal thresholding approach. If $x_i\geq 0.5$, then $x_i=1$ and $x_i=0$,  otherwise. Let $\mathcal{V}_\text{c}$ be the vertex cover found. Then, the independent set $\mathcal{I}=\bar{\mathcal{V}}\backslash \mathcal{V}_\text{c}$ and the clique $\mathcal{C}=\mathcal{I}$. We remove all the vertices of the clique $\mathcal{C}$ from the graph $\mathcal{G}$ and all edges incident on them. Then we find the next dominant clique in the remaining graph.  We find the first $k$ dominant cliques by following  the above procedure. We present the complete procedure in Algorithm \ref{alg:0}.
    			\begin{algorithm}[htbp]
    				\caption{\textit{Symmetry Detection}}\label{alg:0}
    				\begin{algorithmic}[1]
    					\State 
    					\State \textbf{Input}: Graph $\mathcal{G}$, $k$=number of symmetry axes
    					\For{$i\in\{1,2,\ldots,k\}$}
    					\State Construct complement graph $\bar{\mathcal{G}}$ of graph $\mathcal{G}$.
    					\State Find vertex cover $\mathcal{V}_\text{c}$ by solving \ref{eq6} 
    					\State Independent set $\mathcal{I}=\bar{\mathcal{V}}\backslash \mathcal{V}_\text{c}$ 
    					\State Set cluster of MSPPs $\mathcal{P}_i=\{(\mathbf{x}_j,\mathbf{x}_{j^\prime}):j\in\mathcal{I}\}$
    					\State Remove vertices $\mathcal{I}$ from the graph $\mathcal{G}$ and edges incident on them.
    					\EndFor
    					\State  $k$ sets of pairs of mirror symmetric points $\{\mathcal{P}_i\}_{i=1}^k$.  
    				\end{algorithmic}
    			\end{algorithm}
    			
    			Using the pairs of mirror symmetric pixels detected by Algorithm 1, we form the sets $\mathcal{L}$ and $\mathcal{R}$ by picking randomly one pixel of a pair and including it in the set $\mathcal{L}$ and the other pixel in the set $\mathcal{R}$. We further remove any  outlier pairs using the following property of a symmetric function, since the pairs obtained are purely based on geometric constraints. Let the points $\mathbf{x}_i$ and $\mathbf{x}_{i^\prime}$ be mirror reflections each other. Therefore,
    			\begin{eqnarray}
    			\nonumber I(\mathbf{x}_i)=I(\mathbf{x}_{i^\prime})\Rightarrow\nabla_{\mathbf{x}_i} I(\mathbf{x}_i)=\nabla_{\mathbf{x}_i} I(\mathbf{x}_{i^\prime})\\
    			\nonumber \nabla_{\mathbf{x}_i} I(\mathbf{x}_i)=\nabla_{\mathbf{x}_i} I(\mathbf{R}_{ii^\prime}\mathbf{Q}\mathbf{R}_{ii^\prime}^\top \mathbf{x}_i-\mathbf{R}_{ii^\prime}\mathbf{Q}\mathbf{R}_{ii^\prime}^\top\mathbf{t}_{ii^\prime}+\mathbf{t}_{ii^\prime})\\
    			\nabla_{\mathbf{x}_i} I(\mathbf{x}_i)=\mathbf{R}_{ii^\prime}\mathbf{Q}\mathbf{R}_{ii^\prime}^\top\nabla_{\mathbf{x}_{i^\prime}} I(\mathbf{x}_{i^\prime}).
    			\end{eqnarray}
    			We only keep those pairs satisfying $$\nabla_{\mathbf{x}_i} I(\mathbf{x}_i)^\top\mathbf{R}_{ii^\prime}\mathbf{Q}\mathbf{R}_{ii^\prime}^\top\nabla_{\mathbf{x}_{i^\prime}} I(\mathbf{x}_{i^\prime})>1-\epsilon.$$ 
    			Where, $0<\epsilon<1$. 
    			
    			Since the number of pixels, $|\mathcal{E}|$, lying on the edges is very high, it results in huge number $\frac{|\mathcal{E}|(|\mathcal{E}|-1)}{2}$ of pairs. Therefore, we randomly pick pairs and vote in order to reduce the computational complexity. For each edge pixel, we select $h<<|\mathcal{E}|$ pixels which results in a total number of $h|\mathcal{E}|$ pairs. We now show that the probability of selecting the correct mirror reflection pixel of a pixel using the proposed randomization scheme is very high for $h<<|\mathcal{E}|$. Since the symmetry present in the image is approximate symmetry, we consider a pixel to be a mirror reflection even if it is shifted in a square of width $u$ from its ideal location. Now, the probability of selecting the approximate mirror reflection pixel of a pixel under consideration in one attempt is $\frac{u^2}{|\mathcal{E}|-1}$. Therefore, the probability of not selecting in one attempt is $1-\frac{u^2}{|\mathcal{E}|-1}$. Therefore, probability of not selecting the approximate mirror reflection pixel of a pixel under consideration in $h$ attempts is $\big(1-\frac{u^2}{|\mathcal{E}|-1}\big)^h$. Hence, the probability of selecting the approximate mirror reflection pixel of a pixel in $h$ attempts is $1-\big(1-\frac{u^2}{|\mathcal{E}|-1}\big)^h$. For example, if $|\mathcal{E}|=3000$, $h=200$, and $u=5$, this probability is $0.8124$ which is quite high.
    			
    			\begin{figure}[htbp]
    				\centering
    				\stackunder{\epsfig{figure= 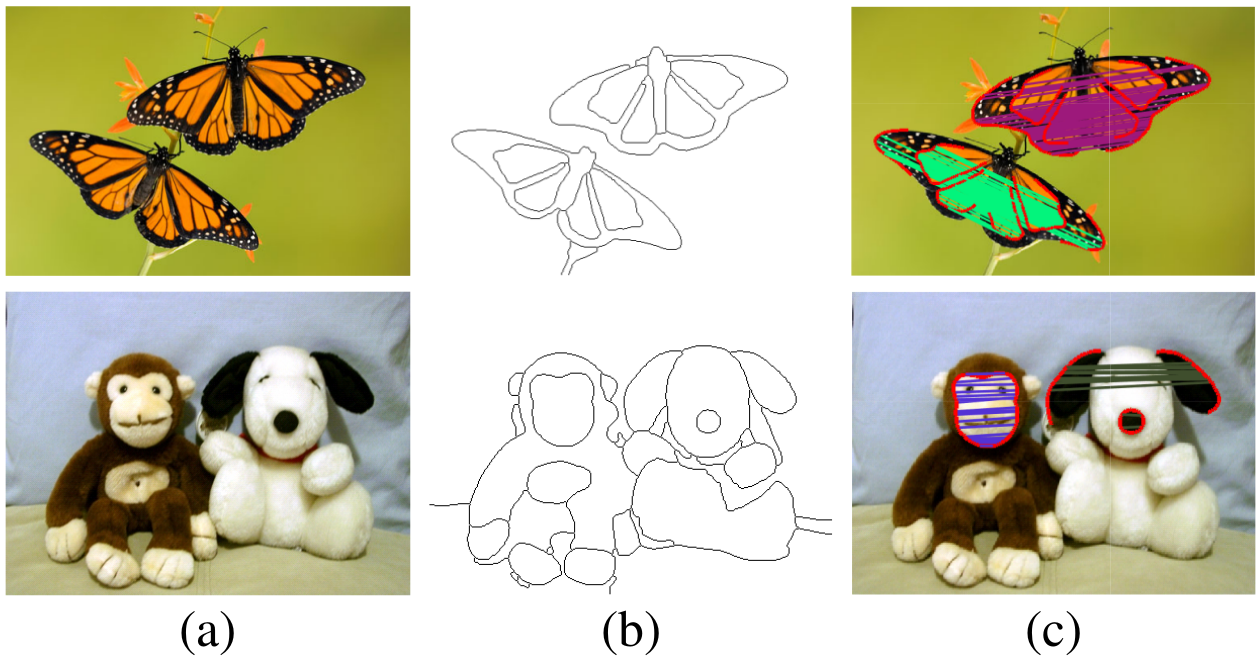,width=1\linewidth}}{}
    				\caption{ (a) Input image, (b) Detected boundaries using the approach \cite{arbelaez2011contour}, and (c) Detected pairs of mirror symmetric pixels. Each set of pairs belonging to the same symmetry axis are colored same.}
    				\label{fig:symmr}
    			\end{figure}
    			In Fig. \ref{fig:symmr}, we present a few results of symmetry detection on the images from the dataset \cite{Funk_2017_ICCV_Workshops}.
    			
    			\subsection{Symmetry Aware SLIC}
    			\label{subsec:slic}
    			In order to preserve the reflection symmetry present in the input image, represented by the sets $\mathcal{L}$ and $\mathcal{R}$, we have to make sure that for each pair $(\mathbf{x}_i,\mathbf{x}_{i^\prime})$ of pixels $\mathbf{x}_i$ and $\mathbf{x}_{i^\prime}$ which are mirror reflections of each other, there should be a pair $(\mathcal{S}_i,\mathcal{S}_{i^\prime})$ of superpixels, $\mathcal{S}_i$ and $\mathcal{S}_{i^\prime}$, which are mirror reflections of each other. We define two superpixels, $\mathcal{S}_i$ and $\mathcal{S}_{i^\prime}$, to be mirror reflections of each other if for each $ \mathbf{x}_j\in\mathcal{S}_i,\exists \mathbf{x}_{j^\prime}\in\mathcal{S}_{i^\prime}$ such that the pixels $\mathbf{x}_{j}$ and $\mathbf{x}_{j^\prime}$ are mirror reflections of each other. We improve and extend the SLIC algorithm proposed in \cite{achanta2012slic} to estimate symmetry aware superpixels. The SLIC algorithm is based on the $k$-means where the goal is to find $k$ center pixels and assignment of each pixel to form $k$ groups or clusters of pixels such that each group contains spatially close and visually similar pixels. In order to preserve the symmetry, we minimize the following objective function, with respect to the centers and the cluster assignments.
    			\begin{eqnarray}
    			\nonumber\text{min }\sum_{i=1}^{k}\sum_{(\mathbf{x},\mathbf{x}^\prime)\in\mathcal{K}_i}\|\mathbf{c}_i-\mathbf{x}\|_2^2+\|\mathbf{c}^\prime_i-\mathbf{x}^\prime\|_2^2\\+\lambda\|(\mathbf{c}_i)-I(\mathbf{x})\|_2^2+\lambda\|I(\mathbf{c}^\prime_i)-I(\mathbf{x}^\prime)\|_2^2
    			\end{eqnarray}
    			
    			Here, $\mathbf{x}^\prime$ represents the mirror image of the pixel $\mathbf{x}$ through the symmetry axis defined by the pairs of symmetric superpixel $\mathcal{S}_i$ and $\mathcal{S}_{i^\prime}$. The set $\mathcal{K}_i$  is the set of pairs of mirror symmetric pixels assigned to the mirror symmetric superpixels  $\mathcal{S}_i$ and $\mathcal{S}_{i^\prime}$.   The cost $\|\mathbf{c}_i-\mathbf{x}\|_2^2+\lambda\|(\mathbf{c}_i)-I(\mathbf{x})\|_2^2$ is similar to the SLIC cost function which make sure that each cluster or superpixel contains spatially close and visually similar pixels. While, the cost $\|\mathbf{c}^\prime_i-\mathbf{x}^\prime\|_2^2+\lambda\|I(\mathbf{c}^\prime_i)-I(\mathbf{x}^\prime)\|_2^2$ ensures that for each pair, if the pixel $\mathbf{x}$ is assigned to the superpixel with center $\mathbf{c}_i$, then the mirror reflection $\mathbf{x}^\prime$ of the pixel $\mathbf{x}$ is assigned to the superpixel with center $\mathbf{c}^\prime_i$ which is the mirror image of the $\mathbf{c}_i$. In order to solve this optimization problem, we follow the general SLIC algorithm. The SLIC algorithm is based on the k-means clustering algorithm. Cluster centers are initialized on the center of equally spaced squares of sizes $s\times s$. In order to update the cluster centers, the distances between a center and all pixels within the square of size $2s\times 2s$ centered at these pixels is computed. Each pixel is assigned to the nearest center. Then the center is updated using its new neighboring pixels by taking the average location and the average color. This process is continued till convergence. Here $\lambda$ is the compactness factor and generally chosen in the range $[1, 40]$ \cite{achanta2012slic}. Higher values of $\lambda$ result in compact superpixels and poor boundary adherence and lower values $\lambda$ result in poor compactness of superpixels and better adherence to boundaries. \\
    			\begin{figure}[htbp]
    				\centering
    				\stackunder{\epsfig{figure= 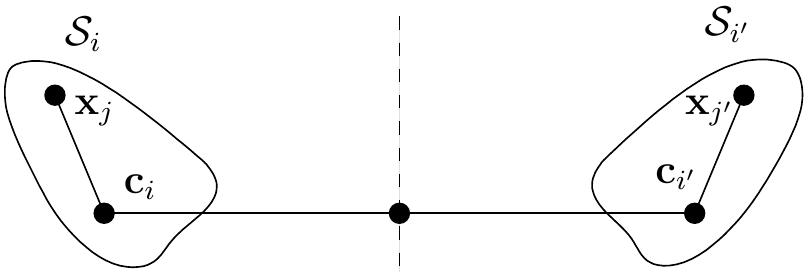,width=0.9\linewidth}}{}
    				\caption{ The symmetric assignment: if the superpixels $\mathcal{S}_i$ and $\mathcal{S}_{i^\prime}$ are mirror reflections of each other, pixels $\mathbf{x}_i$ and $\mathbf{x}_{i^\prime}$ are mirror reflections of each other, and the pixel $\mathbf{x}_j$ is assigned to the center $\mathbf{c}_i$, then we assign the pixel $\mathbf{x}_{j^\prime}$ to the center $\mathbf{c}_{i^\prime}$.}
    				\label{fig:mspp1}
    			\end{figure}
    			\textbf{Symmetric Initialization.} In order to preserve the reflection symmetry represented in the sets $\mathcal{L}$ and $\mathcal{R}$, we have to make sure that for a pair of pixels which are mirror reflections of each other, there should be a corresponding pair of superpixels which are mirror reflections of each other. Let the pixels $\mathbf{x}_i\in\mathcal{L}$ and $\mathbf{x}_{i^\prime}\in\mathcal{R}$ be mirror reflections of each other. We initialize the centers $\mathbf{c}_i$ and $\mathbf{c}_{i^\prime}$ of two superpixels $\mathcal{S}_i$ and $\mathcal{S}_{i^\prime}$  at the pixels $\mathbf{x}_i$ and $\mathbf{x}_{i^\prime}$. We observe that the symmetric object present in the image might not cover the full image and therefore in the non-symmetric region, we follow the same initialization strategy as used in SLIC. We first find the convex hull, $\mathcal{C}$, of the set $\mathcal{L}\cup\mathcal{R}$ which represent the symmetric region.
    		Now, in the non-symmetric region $\{\mathcal{W}\times\mathcal{H}\}\setminus\mathcal{C}$, we initialize the centers at the centers of equally spaced squares and in the region $\mathcal{C}$, we do the symmetric initialization. We observe that the reflection symmetric pixels obtained lie on the edges. Therefore, according to \cite{achanta2012slic}, it is an unstable initialization. However, we observe that these pairs of reflection symmetric pixels exhibit high accuracy. Therefore, we transfer each pair to a new location such that the image gradient at both the pixels of the new pair is minimum in the local vicinity and are mirror reflections of each other.  
    			\\
    			\textbf{Number of superpixels.} If we want $k$ superpixels, then we partition the image into square windows of sides equal to $\sqrt{\frac{wh}{k}}$. If the number of pairs of mirror symmetric points $\mid\mathcal{L}\mid$ in the symmetric region $\mathcal{C}$ is greater than $(k-s)$, then we randomly select $(k-s)$ pairs from the $\mid\mathcal{L}\mid$ pairs. Here, $s$ is the number square windows in the non-symmetric region $\{\mathcal{W}\times\mathcal{H}\}\setminus\mathcal{C}$. If the number pairs of mirror symmetric points $\mid\mathcal{L}\mid$ in the symmetric region $\mathcal{C}$ is less than $(k-s)$, then we randomly select $(k-s-\mid\mathcal{L}\mid)$ points in the symmetric region and  reflect them using the symmetry axis defined by their nearest pair.\\
    			\textbf{Symmetric Assignment.} We propose an assignment strategy in order to achieve pairs of reflection symmetric superpixels with equal areas and similar boundaries. We assign pixels to their nearest centers such that each pair of superpixels $(\mathcal{S}_i, \mathcal{S}_{i^\prime})$ remains mirror reflection of each other in all iterations. Let us consider the centers $\mathbf{c}_i$ and $\mathbf{c}_{i^\prime}$ of two superpixels, $\mathcal{S}_i$ and $\mathcal{S}_{i^\prime}$, which are mirror reflections of each other. Let $\mathbf{x}_j$ be a pixel inside the square of size $2s\times 2s$ around the center $\mathbf{c}_i$. If the nearest center to the pixel $\mathbf{x}_j$ is $\mathbf{c}_i$, then we assign the center $\mathbf{c}_{i^\prime}$ as the nearest center to the pixel $\mathbf{x}_{j^\prime}=\mathbf{R}_{ii^\prime}\mathbf{Q}\mathbf{R}_{ii^\prime}^\top(\mathbf{x}_j-\mathbf{t}_{ii^\prime})+\mathbf{t}_{ii^\prime}$. Here, $\mathbf{t}_{ii^\prime}=\frac{\mathbf{c}_i+\mathbf{c}_{i^\prime}}{2}$ and $\mathbf{R}_{ii^\prime}$ is the rotation matrix with angle $\theta_{ii^\prime}$ equal to the slope of the line passing through the pixels $\mathbf{c}_i$ and $\mathbf{c}_{i^\prime}$. Fig. \ref{fig:mspp1} graphically illustrates this concept. We prove that, using this assignment strategy, a pair of reflection symmetric superpixels remains a pair of reflection symmetric superpixels after one iteration.\\
    			\textbf{Claim 1.} \textit{Let $\mathcal{S}_i^t$ and $\mathcal{S}_{i^\prime}^t$ be two superpixels which are mirror reflections of each other at the iteration $t$. Let $\mathbf{c}_i^t$ and $\mathbf{c}_{i^\prime}^t$ be their centers respectively. Then, at the iteration $t+1$, the updated superpixels $\mathcal{S}_i^{t+1}$ and $\mathcal{S}_{i^\prime}^{t+1}$ will also be mirror reflections of each other.}\\
    			\textbf{Proof.} Using assignment strategy, if we assign $\mathbf{x}_i$ to  $\mathcal{S}_i$, then we assign the pixel $\mathbf{R}_{ii^\prime}\mathbf{Q}\mathbf{R}_{ii^\prime}^\top\mathbf{x}_j-\mathbf{R}_{ii^\prime}\mathbf{Q}\mathbf{R}_{ii^\prime}^\top\mathbf{t}_{ii^\prime}+\mathbf{t}_{ii^\prime}$ to  $\mathcal{S}_{i^\prime}$. Let us assume that we assign $n_{i}$ pixels to the superpixel $\mathcal{S}_i$, and $\mathcal{J}_i=\{i_1,i_2,\ldots,i_{n_i}\}$, $\mathcal{J}_{i^\prime}=\{i^\prime_1,i^\prime_2,\ldots,i^\prime_{n_{i}}\}$ be the sets of indices of pixels belonging to the superpixels $\mathcal{S}_i$ and $\mathcal{S}_{i^\prime}$, respectively. The center of the superpixel $\mathcal{S}_i$ is $\mathbf{c}^{t+1}_i$=$\frac{1}{n_i}\sum_{j\in\mathcal{J}_i}\mathbf{x}_{j}$. Now, the center of the superpixel $\mathcal{S}_{i^\prime}$ is 
    			
    			$$\mathbf{c}_{i^\prime}^{t+1}=\frac{1}{n_i}\sum_{j^\prime\in\mathcal{J}_{i}}\mathbf{x}_{j^\prime}=
    			$$
    			$$\frac{1}{n_i}\sum_{j\in\mathcal{J}_{i}}\mathbf{R}_{ii^\prime}\mathbf{Q}\mathbf{R}_{ii^\prime}^\top\mathbf{x}_j-\frac{1}{n_i}\sum_{j\in\mathcal{J}_{i^\prime}}\mathbf{R}_{ii^\prime}\mathbf{Q}\mathbf{R}_{ii^\prime}^\top\mathbf{t}_{ii^\prime}+\frac{1}{n_i}\sum_{j\in\mathcal{J}_{i^\prime}}\mathbf{t}_{ii^\prime}$$
    			$$=\mathbf{R}_{ii^\prime}\mathbf{Q}\mathbf{R}_{ii^\prime}^\top\frac{1}{n_i}\sum_{j\in\mathcal{J}_{i}}\mathbf{x}_j-\mathbf{R}_{ii^\prime}\mathbf{Q}\mathbf{R}_{ii^\prime}^\top\mathbf{t}_{ii^\prime}+\mathbf{t}_{ii^\prime}$$
    			$$=\mathbf{R}_{ii^\prime}\mathbf{Q}\mathbf{R}_{ii^\prime}^\top\mathbf{c}^{t+1}_i-\mathbf{R}_{ii^\prime}\mathbf{Q}\mathbf{R}_{ii^\prime}^\top\mathbf{t}_{ii^\prime}+\mathbf{t}_{ii^\prime}.
    			$$
    			Therefore, the center of superpixels at the iteration $t+1$ and the updated superpixels $\mathcal{S}_i^{t+1}$ and $\mathcal{S}_{i^\prime}^{t+1}$ will also be mirror reflections of each other. $\square$
    			
    			We further observe that the centers of mirror symmetric superpixels follow the curves that are mirror reflections of each other. It is easy to prove this claim from the Claim 1. In Algorithm \ref{alg:1}, we present all the steps involved in the proposed \textit{SymmSLIC} algorithm.
    				\begin{algorithm}[htbp]
    					\caption{\textit{SymmSLIC}}\label{alg:1}
    					\begin{algorithmic}[1]
    						\State \textbf{Input}: Image $I$, number of superpixels $k$, and parameter $\lambda$.
    						\State Solution:
    						\State Initialize the label matrix $\mathbf{L}(\mathbf{x}_i)=-1,\text{ and the distance matrix }  \mathbf{D}(\mathbf{x}_i)=\infty,\;\forall\mathbf{x}_i\in\mathcal{W}\times\mathcal{H}$.
    						\State Determine the sets $\mathcal{L}$ and $\mathcal{R}$ as discussed in Section \ref{subsec:RA}.
    						\State Initialize the cluster centers as discussed in Section \ref{subsec:slic}.
    						\While{not converged}
    						\For{ each cluster center $\mathbf{c}_i$ in $\mathcal{L}$}
    						\State Determine $\mathbf{R}_{ii^\prime}$ and $\mathbf{t}_{ii^\prime}$ using $\mathbf{c}_i$ and $\mathbf{c}_{i^\prime}$.
    						\For{each pixel $\mathbf{x}_j$ in the $2s\times 2s$ square around  $\mathbf{c}_i$}
    						\State Compute the distance $d(\mathbf{x}_j,\mathbf{c}_i)$ between $\mathbf{c}_i$ and  $\mathbf{x}_j$.
    						\If {$d(\mathbf{x}_j,\mathbf{c}_i)<\mathbf{D}(\mathbf{x}_j)$}
    						\State$\mathbf{D}(\mathbf{x}_j)\gets d(\mathbf{x}_j,\mathbf{c}_i)$ and $\mathbf{L}(\mathbf{x}_j)\gets i$
    						\State$\mathbf{x}_{j^\prime}\gets\mathbf{R}_{ii^\prime}\mathbf{Q}\mathbf{R}_{ii^\prime}^\top\mathbf{x}_j-\mathbf{R}_{ii^\prime}\mathbf{Q}\mathbf{R}_{ii^\prime}^\top\mathbf{t}_{ii^\prime}+\mathbf{t}_{ii^\prime}$
    						\State$\mathbf{D}(\mathbf{x}_{j^\prime})\gets d(\mathbf{x}_{j^\prime},\mathbf{c}_{i^\prime})$ and $\mathbf{L}(\mathbf{x}_{j^\prime})\gets i^\prime$
    						\EndIf
    						
    						\EndFor
    						\EndFor
    						\For{ each cluster center $\mathbf{c}_{i^\prime}$ in $\mathcal{R}$}
    						\State Determine $\mathbf{R}_{ii^\prime}$ and $\mathbf{t}_{ii^\prime}$ using $\mathbf{c}_i$ and $\mathbf{c}_{i^\prime}$.
    						\For{each pixel $\mathbf{x}_{j^\prime}$ in the $2s\times 2s$ square around  $\mathbf{c}_{i^\prime}$}
    						\State Compute the distance $d(\mathbf{x}_{j^\prime},\mathbf{c}_{i^\prime})$ between $\mathbf{c}_{i^\prime}$ and  $\mathbf{x}_{j^\prime}$.
    						\If {$d(\mathbf{x}_{j^\prime},\mathbf{c}_{i^\prime})<\mathbf{D}(\mathbf{x}_{j^\prime})$}
    						\State$\mathbf{D}(\mathbf{x}_{j^\prime})\gets d(\mathbf{x}_{j^\prime},\mathbf{c}_{i^\prime})$ and $\mathbf{L}(\mathbf{x}_{j^\prime})\gets i^\prime$
    						\State$\mathbf{x}_{j}\gets\mathbf{R}_{ii^\prime}\mathbf{Q}\mathbf{R}_{ii^\prime}^\top\mathbf{x}_{j^\prime}-\mathbf{R}_{ii^\prime}\mathbf{Q}\mathbf{R}_{ii^\prime}^\top\mathbf{t}_{ii^\prime}+\mathbf{t}_{ii^\prime}$
    						\State$\mathbf{D}(\mathbf{x}_j)\gets d(\mathbf{x}_j,\mathbf{c}_i)$ and $\mathbf{L}(\mathbf{x}_j)\gets i$
    						\EndIf
    						\EndFor
    						\EndFor
    						\For{all the centers in the region $\big(\{\mathcal{W}\times\mathcal{H}\}\setminus\text{convexhull}(\mathcal{L}\cup\mathcal{R})\big)$}
    						\State Perform SLIC.
    						\EndFor
    						\State Update the cluster centers.
    						\EndWhile
    					\end{algorithmic}
    				\end{algorithm} 
    			\begin{figure*}[htbp]
    				\centering
    				\stackunder{\epsfig{figure= 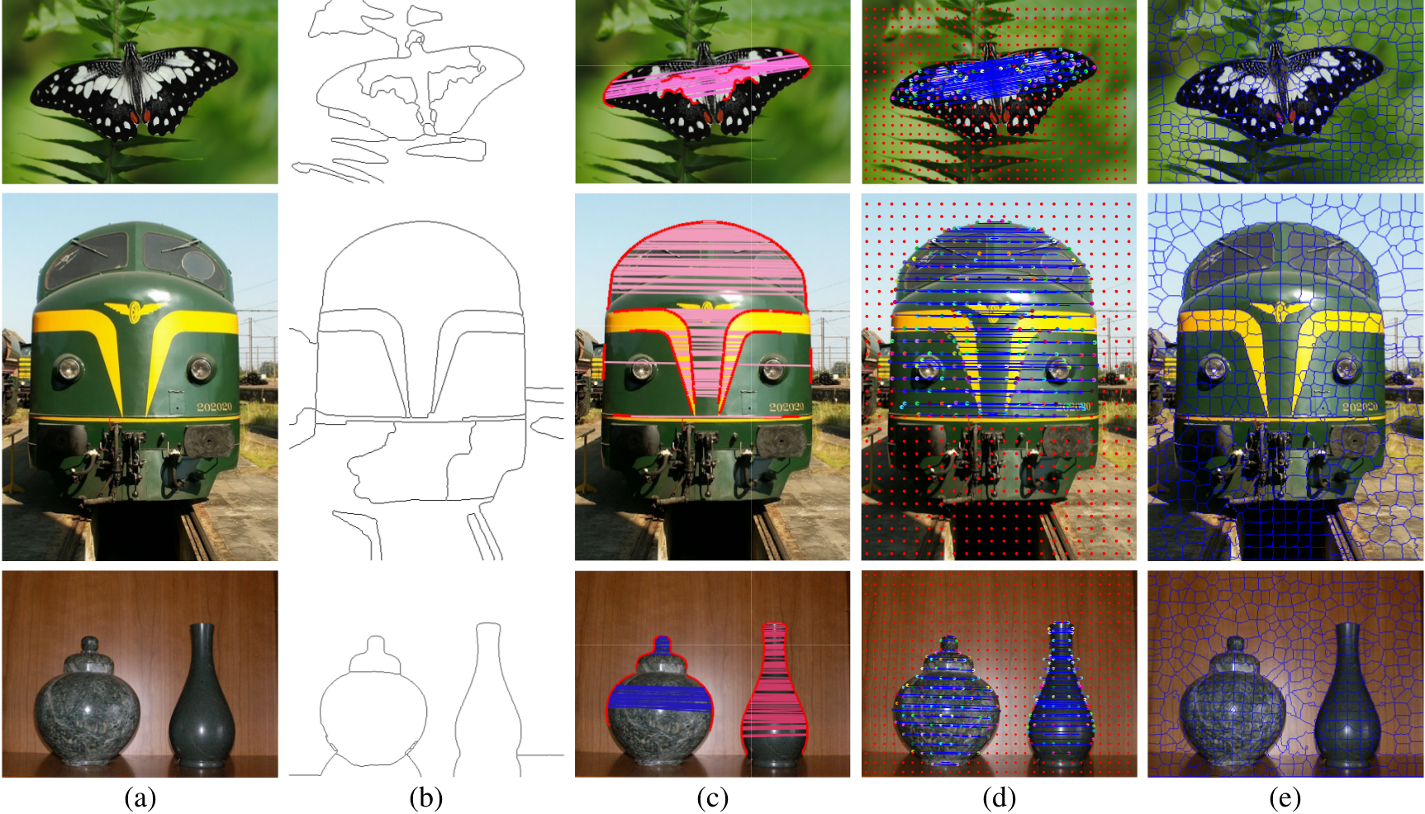,width=1\linewidth}}{(e)}    
    				\caption{(a) Input image $I$, (b), detected edges, (c) the pairs of mirror symmetric points ($\mathcal{L}$ and $\mathcal{R}$), (d) the initialized centers, the red centers are in the non-symmetric regions and the green centers are in the symmetric regions, and (e) the symmetric superpixel segmentation with the pairs of mirror symmetric superpixels.}
    				\label{fig:all}
    			\end{figure*}

    			\begin{figure*}
    				\centering
    				\stackunder{\epsfig{figure= 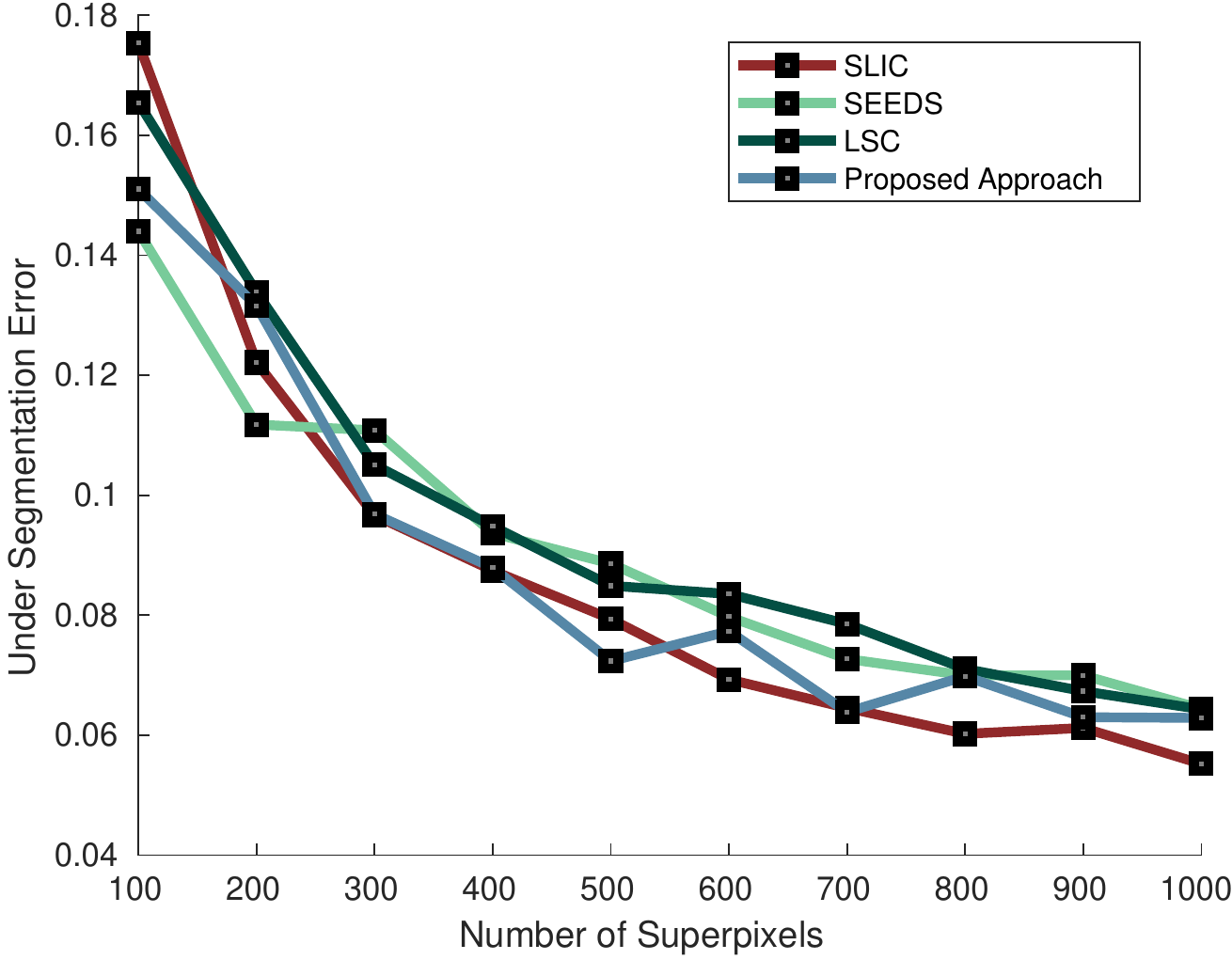,width=0.32\linewidth}}{(a)}
    				\stackunder{\epsfig{figure= 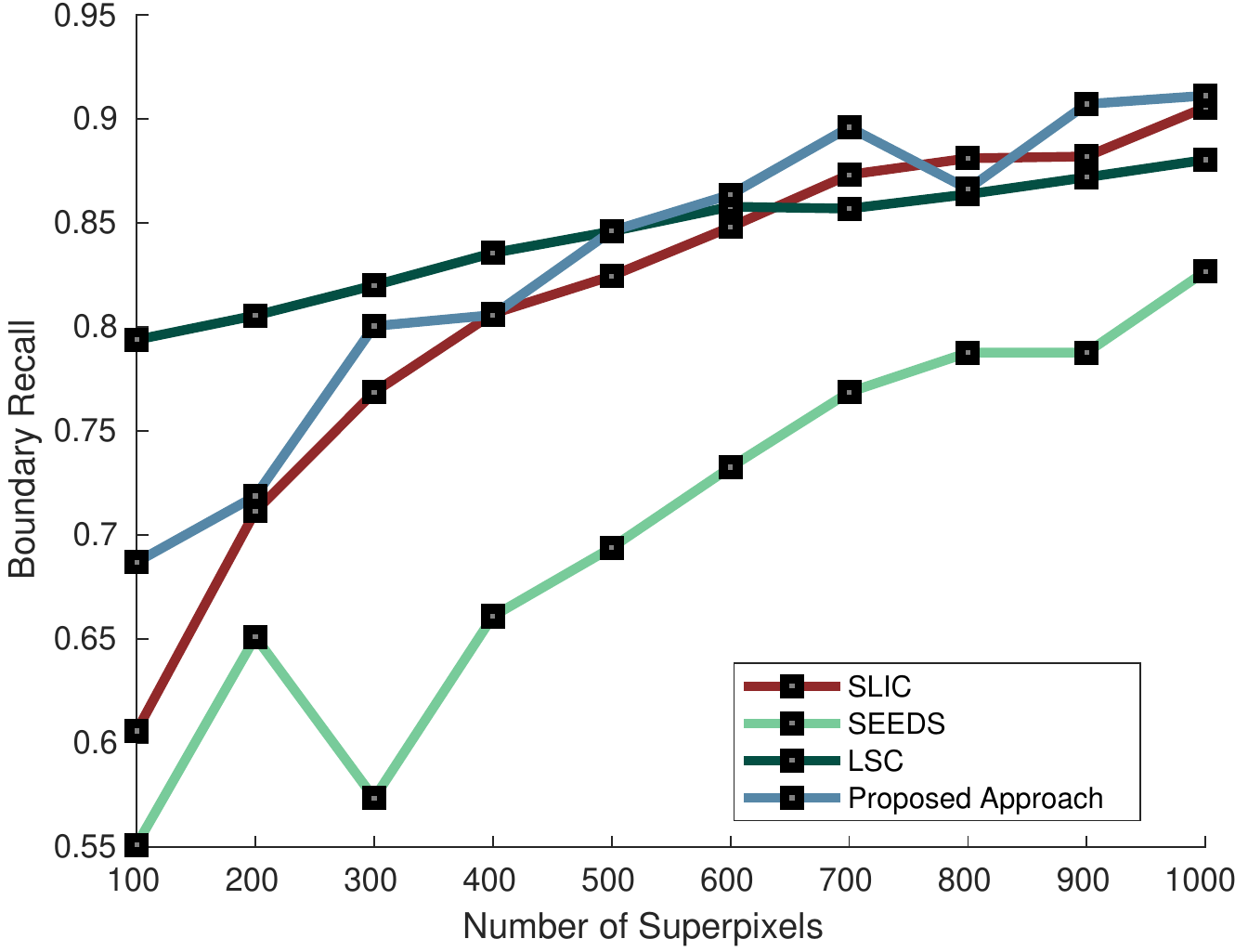,width=0.32\linewidth}}{(b)}
    				\stackunder{\epsfig{figure= 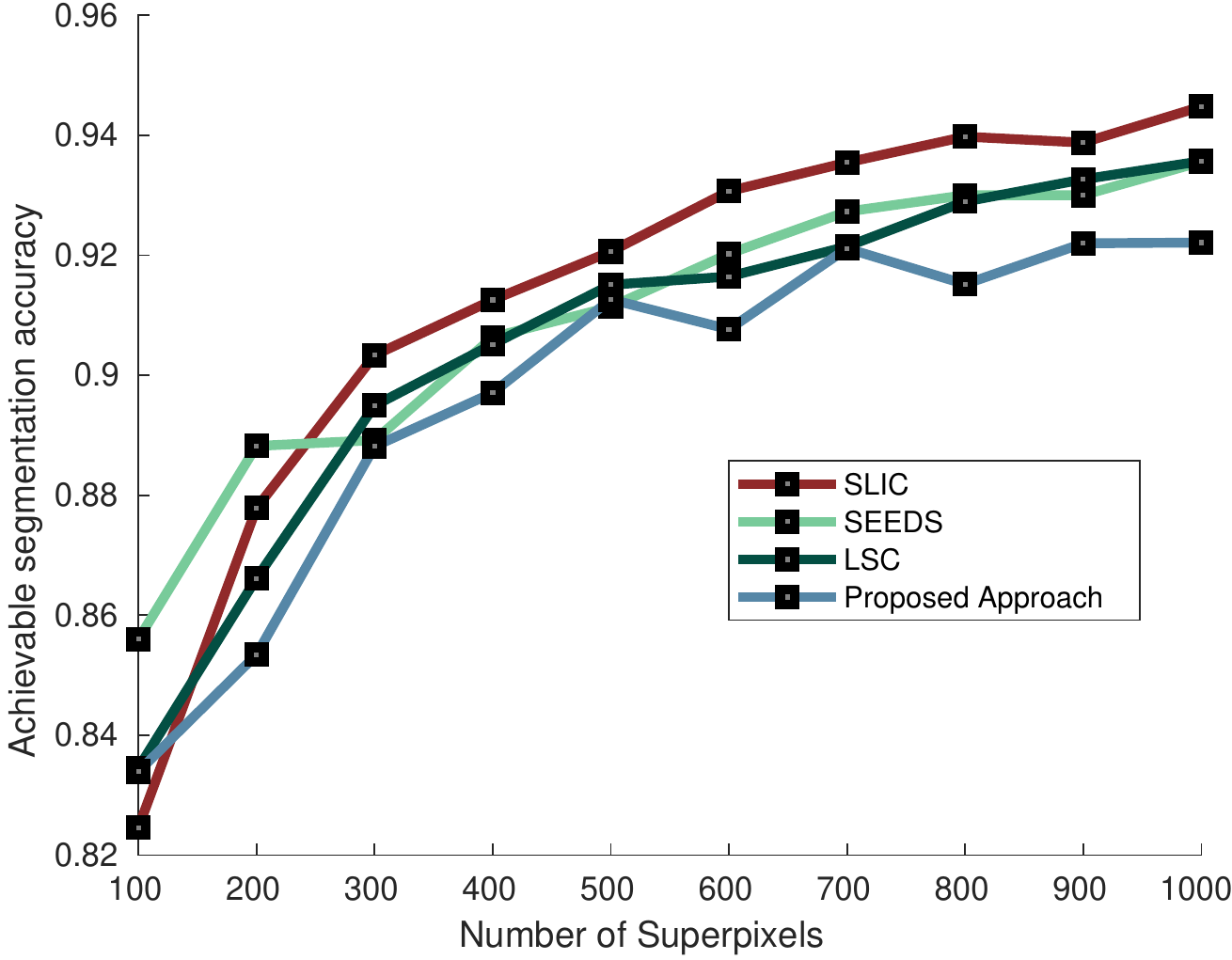,width=0.32\linewidth}}{(c)}
    				\caption{(a) Under segmentation error, (b) Boundary recall, and (c) Achievable segmentation accuracy vs the number of superpixels plots.}
    				\label{fig:ev}
    			\end{figure*}
    			\begin{figure}[!h]
    				\centering
    				\stackunder{\epsfig{figure= 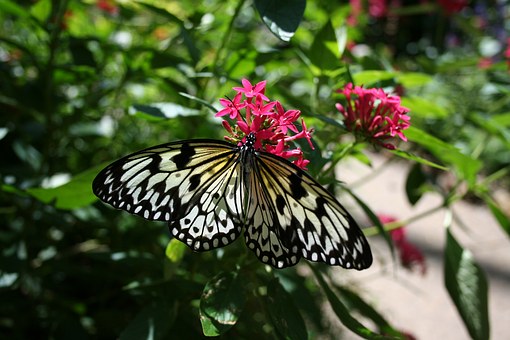,width=0.32\linewidth}}{}
    				\stackunder{\epsfig{figure= 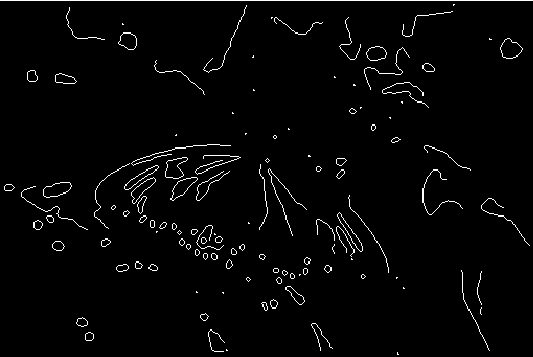,width=0.32\linewidth}}{}
    				\stackunder{\epsfig{figure= 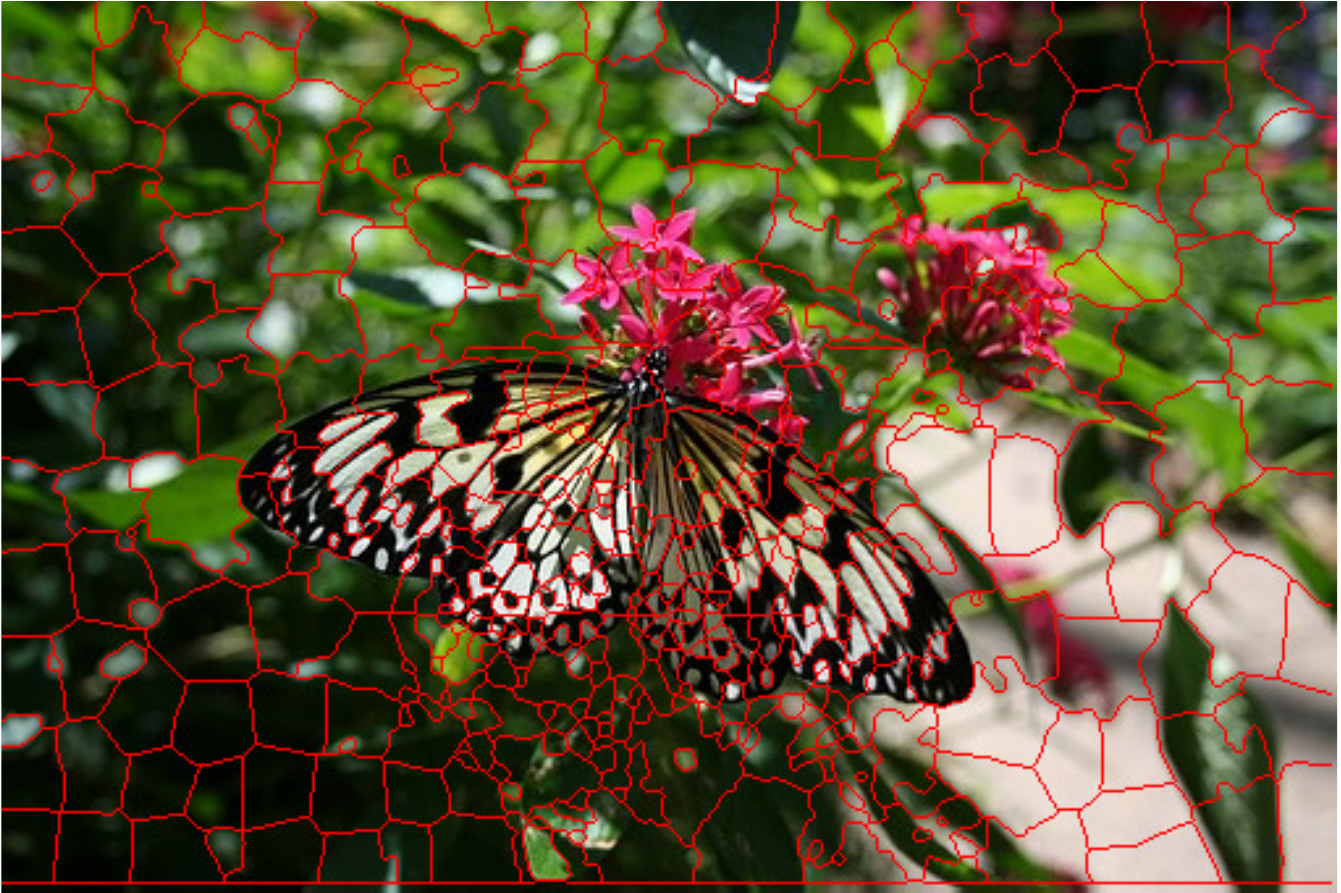,width=0.32\linewidth}}{}
    				\caption{A failure case: Input, Edges, and SymmSLIC}
    				\label{fig:fail}
    			\end{figure} 
    			\section{Results and Evaluation}
    			\label{sec:RD}
    			In Figure \ref{fig:all}, we show the major steps of the proposed approach using a few example images. We observe that the symmetry present at the pixel level, as shown in Figure \ref{fig:all} (c), is well preserved at the superpixel level, as shown in Figure \ref{fig:all} (e).  In Figure \ref{fig:fail}, we present a failure case. The SymmSLIC fails due to the improper detection of the edges across the symmetry axis. However, we observe that the resulting over-segmentation is very similar to the SLIC superpixels.  
    			
    			For the quantitative evaluation, we measure the boundary recall and the under-segmentation error for the proposed approach and the approaches \cite{achanta2012slic}, \cite{li2015superpixel}, and \cite{van2015seeds} on the dataset BSDS500 \cite{martin2001database}. We do evaluation only on the images containing symmetric objects from this dataset. In order to measure the performance, we use  three metrics: under segmentation error \cite{levinshtein2009turbopixels}, boundary recall \cite{martin2004learning}, and achievable segmentation accuracy \cite{liu2011entropy}. The under segmentation error measures the leakage of estimated superpixels. The higher values of boundary recall represent better adherence to the ground truth boundaries. The achievable segmentation accuracy measures the object segmentation accuracy which can be achieved through the estimated superpixels.  In Figure \ref{fig:ev}, we plot the under segmentation error, boundary recall, and achievable segmentation accuracy as the function of number of superpixels for SLIC \cite{achanta2012slic}, LSC \cite{li2015superpixel}, SEEDS \cite{van2015seeds} and the proposed approach. We observe that the performance of the proposed method is comparable to that of the SLIC \cite{achanta2012slic}, since SymmSLIC works similar to SLIC algorithm and on average almost 60 \% region of the images is non-symmetric. Therefore, in these regions, only classical SLIC is performed.   
    			
\begin{figure*}[htbp]
						\epsfig{figure= 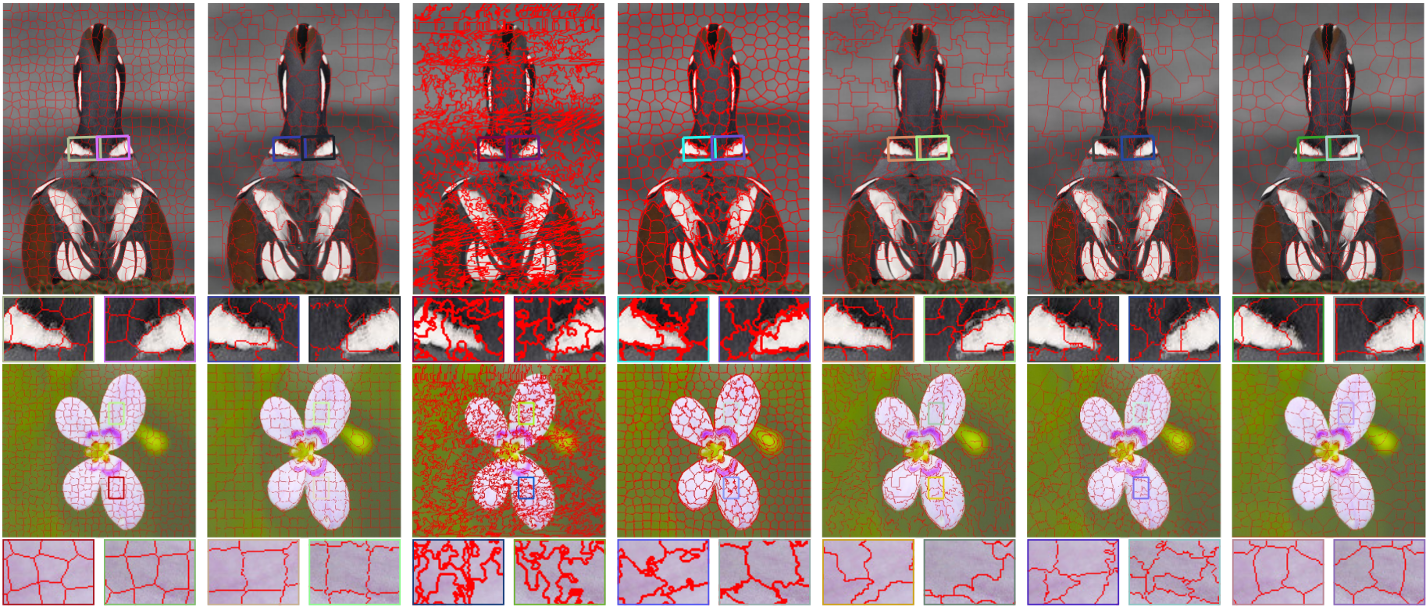,width=1\linewidth}
						\epsfig{figure= 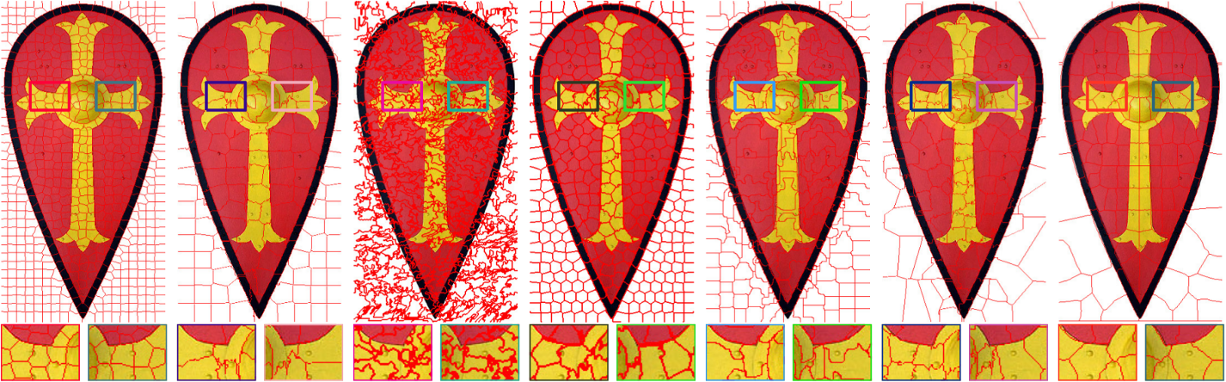,width=1\linewidth}
						\epsfig{figure= 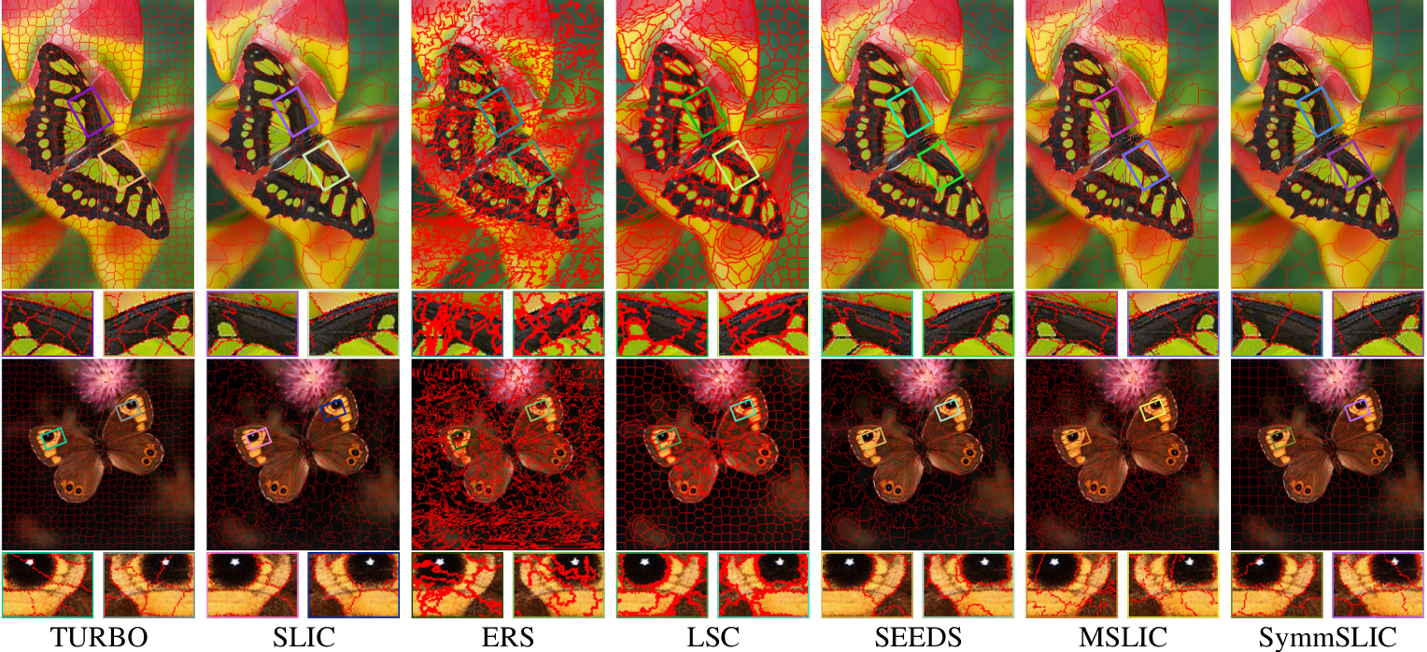,width=1\linewidth}                                                 
				\caption{Results for the approaches  TURBO \cite{levinshtein2009turbopixels}, SLIC \cite{achanta2012slic}, ERS \cite{liu2011entropy}, LSC \cite{li2015superpixel}, SEEDS \cite{van2015seeds}, MSLIC \cite{liu2016manifold}, and the proposed method. For each image, we show two zoomed-in mirror symmetric windows to visualize whether the superpixels are mirror reflections of each other or not. Window border represents its location in the image. The tilted windows are aligned horizontally.}
				\label{fig:results}
			\end{figure*}

    			In Figure \ref{fig:results}, we show the results of the proposed approach. We compare the results of the proposed approach to the results of the methods TURBO \cite{levinshtein2009turbopixels}, SLIC \cite{achanta2012slic}, ERS \cite{liu2011entropy}, LSC \cite{li2015superpixel}, SEEDS \cite{van2015seeds}, and MSLIC \cite{liu2016manifold}.  In each image, we zoom two windows which are mirror reflections of each other. In odd-numbered rows, we show the results obtained on images for all the methods. In the even-numbered rows, we show two zoomed-in mirror symmetric windows from the images in the odd-numbered rows. There does not exist any dataset in which, for an image containing symmetric objects, the ground truth pairs of reflection symmetric pairs of superpixels are present. Therefore, we only report the results obtained through all the methods on images containing symmetric objects. We choose the length, $p$,  of the curve $c_{\mathbf{x}}(\alpha)$ to be equal to 64 pixels, the threshold $\epsilon=0.2$, and the variable $\lambda\in[1,40]$. We observe that using the proposed algorithm \textit{SymmSLIC}, we are able to generate pairs of mirror symmetric superpixels in the symmetric regions. In some cases, our algorithm partitions a perceptually uniform region into many superpixels and due to the symmetric assignment strategy, we achieve a similar segmentation in the mirror symmetric counterpart. 
    			We implemented SymmSLIC in MATLAB on a 2.90GHz$\times$4, 8GB RAM machine. The average time is $\sim 10s$ for an image of size $640\times480$ for 500 superpixels including the detection of pairs of pixels which are mirror reflections of each other.
    			
    			\section{Applications}
    			
    			\subsection{Symmetry Axis Detection}
\begin{figure}[!h]
    				\centering
    				\epsfig{figure= 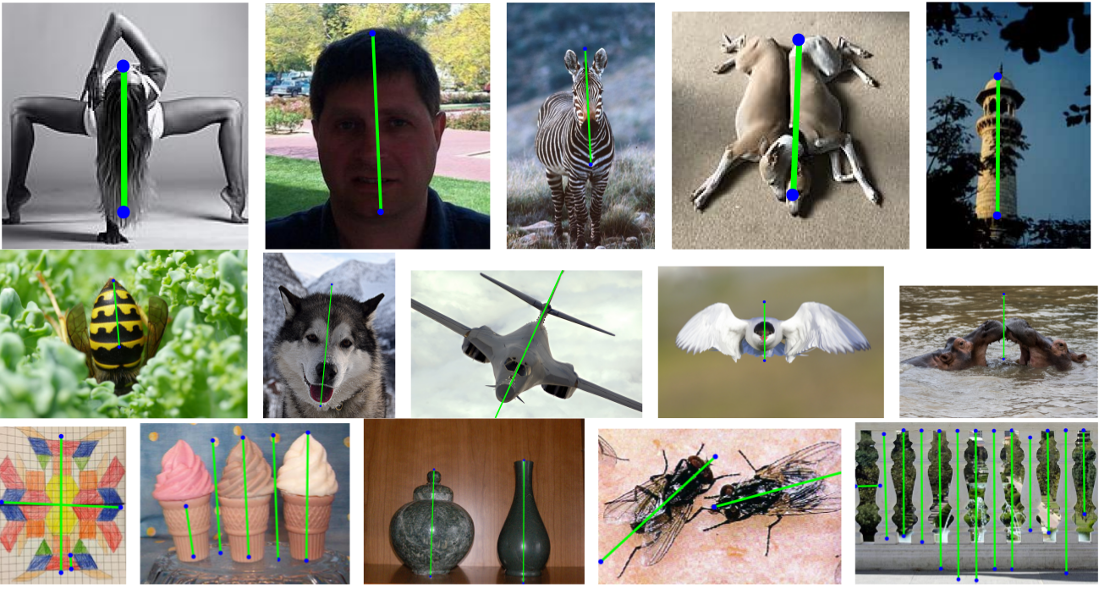,width=1\linewidth}
    				\caption{Symmetry axes detection results on the dataset \cite{Funk_2017_ICCV_Workshops}. First two rows: single symmetry axis, and last row: multiple symmetry axes.}
    				\label{fig:sas}
    			\end{figure}

    			We use the detected pairs mirror symmetric pixels in the Section \ref{subsec:RA} to detect the symmetry axes of the reflective symmetric objects present in the input image. We represent the detected pairs of mirror symmetric pixels as the collection of sets $\{\mathcal{P}_i\}_{i=1}^{k}$ such that each set $\mathcal{P}_i$ contains pairs of mirror symmetric pixels which are symmetric about a same axis. Each pair $(\mathbf{x}_j,\mathbf{x}_{j^\prime})\in\mathcal{P}_i$ defines its own symmetry axis which the line passing through the point $\frac{\mathbf{x}_j+\mathbf{x}_{j^\prime}}{2}$ and is perpendicular to the vector $\mathbf{x}_j-\mathbf{x}_{j^\prime}$. Since all the pairs in the set $\mathcal{P}_i$ belongs to a same symmetry axis the symmetry axes defined all the pairs should be similar. Hence, the  best symmetry axis which is close to all the candidate symmetry axes is the average line passing through the point $\sum_{(\mathbf{x}_j,\mathbf{x}_{j^\prime})\in\mathcal{P}_i}\frac{\mathbf{x}_j+\mathbf{x}_{j^\prime}}{2|\mathcal{P}_i|}$ and perpendicular to the vector $\sum_{(\mathbf{x}_j,\mathbf{x}_{j^\prime})\in\mathcal{P}_i}(\mathbf{x}_j-\mathbf{x}_{j^\prime})$ as proposed in \cite{nagar2017reflection}. In Fig. \ref{fig:sas}, we show the detected symmetry axes on  few images from the dataset \cite{Funk_2017_ICCV_Workshops}. We compare our method using F-score with the methods \cite{michaelsen2017hierarchical}, \cite{elawady2017wavelet}, \cite{guerrini2017innerspec}, \cite{cicconet2017finding}, \cite{atadjanov2016reflection}, \cite{loy2006detecting}, for single symmetry axis detection on the dataset in \cite{Funk_2017_ICCV_Workshops}. In the TABLE \ref{tab1}, we show the $F$-score for all the methods. We observe that the proposed approach is able to achieve the state-of-the-art performance on the dataset in \cite{Funk_2017_ICCV_Workshops}. Here, $F\text{-score}=\frac{2tp}{2tp+fp+fn}$, $tp=$ number of correctly detected axes, $fp=$ number of incorrectly detected axes, and $fn=$ number missed ground-truth axes.  
    			
    			\begin{table}
    				\begin{center}
    					\caption{$F$-score for the methods \cite{michaelsen2017hierarchical}, \cite{elawady2017wavelet}, \cite{guerrini2017innerspec}, \cite{cicconet2017finding}, \cite{atadjanov2016reflection}, \cite{loy2006detecting}, and proposed method on the images of the dataset \cite{Funk_2017_ICCV_Workshops}.}
    					\label{tab1}
    					\begin{tabular}{|c|c|c|c|c|c|c|c|}
    						\hline
    						&\cite{michaelsen2017hierarchical}&\cite{elawady2017wavelet}&\cite{guerrini2017innerspec}&\cite{cicconet2017finding}&\cite{atadjanov2016reflection}&\cite{loy2006detecting}&Ours\\\hline
    						Single&0.27&0.40&0.16&0.38&0.52&0.48&\textbf{0.61}\\\hline
    						Multiple&0.12&0.22&-&-&0.21&\textbf{0.30}&\textbf{0.30}\\\hline
    					\end{tabular}
    				\end{center}
    			\end{table}

    			\subsection{ Unsupervised Symmetric Object Segmentation}
    			\label{sec:5}
    			Object segmentation is a challenging problem which is generally performed either with user interaction and graph cuts (\cite{rother2004grabcut}, \cite{fu2014symmetry}, \cite{tang2015secrets}, \cite{tang2013grabcut}) or with supervised learning \cite{hariharan2014simultaneous}. Our approach also differs from \cite{levinshtein2013multiscale} in the sense that they perform local symmetry grouping whereas we perform global symmetry grouping.  We would like to demonstrate how object segmentation can be performed in an unsupervised manner using SymmSLIC. This section is meant to illustrate as to how the algorithm developed in this paper can be used to solve this classic computer vision application. The application is however limited to images containing objects exhibiting reflection symmetry. 
    			
\begin{figure*}[!h]
    				\centering
    				\stackunder{\epsfig{figure= 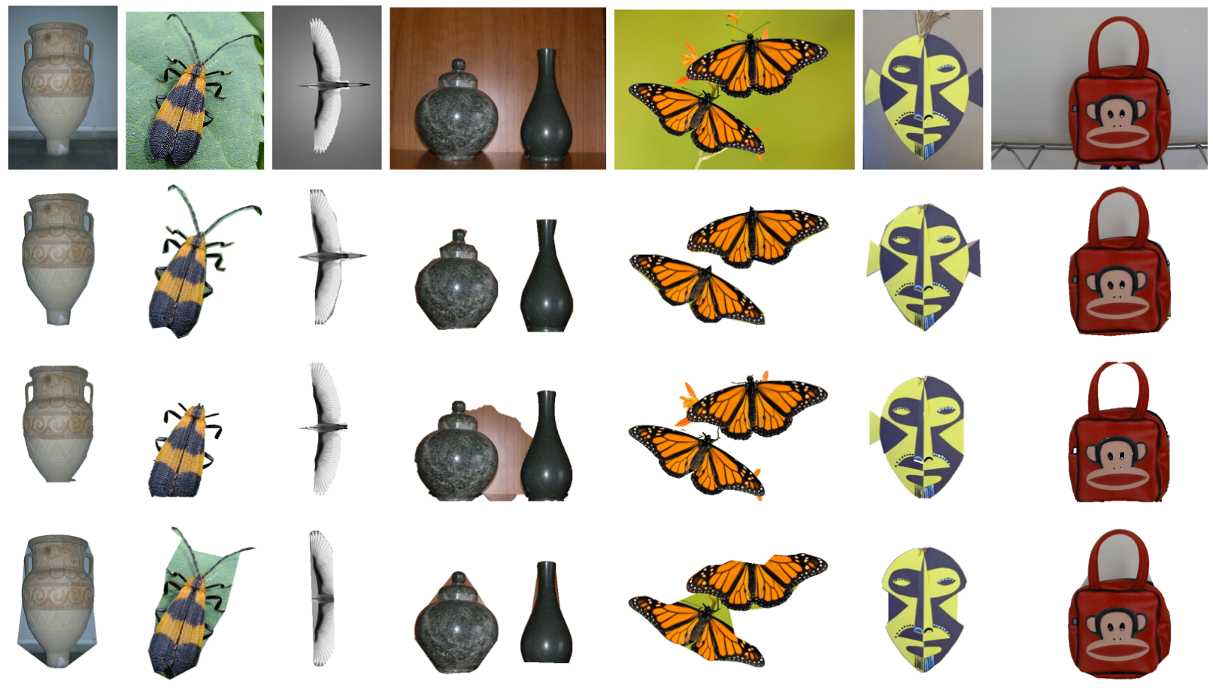,width=01\linewidth}}{}
    				\caption{Row 1: Input images, Row 2: the ground truth object segmentation, Row 3:  grab cut in one cut \cite{tang2013grabcut} segmentation which requires a input mask around the object, and Row 4: unsupervised symmetric objects segmentation using proposed approach which does not require any input mask.}
    				\label{fig:seg}
    			\end{figure*}

    			We use the  SymmSLIC superpixels to segment a symmetric object. This approach is clearly an unsupervised object segmentation approach. The proposed segment is the area, $\cup_{i\in \mathcal{I}}\mathcal{S}_i\cup\mathcal{S}_{i^\prime}$, occupied by the pairs $(\mathcal{S}_i,\mathcal{S}_{i^\prime})$ of the superpixels $\mathcal{S}_i$ and $\mathcal{S}_{i^\prime}$, which are mirror reflections of each other. Here $\mathcal{I}$ is the set of indices of the pairs of mirror symmetric superpixels. We compare our method with the state-of-the-art interactive method \cite{tang2013grabcut} on this challenge  dataset \cite{Funk_2017_ICCV_Workshops}. The method in \cite{tang2013grabcut} assumes that the bounding box around the symmetric object is given. Whereas, our method does not require any such user interaction. In this dataset \cite{Funk_2017_ICCV_Workshops}, each 2D reflection symmetric image contains a symmetric object. We manually created the ground truth segmentations. We compute the error rate defined as the ratio of the misclassified pixels to the total number of pixels. The averaged error rate on all the images from \cite{Funk_2017_ICCV_Workshops} for the method in \cite{tang2013grabcut} is 0.15 and for the proposed approach is 0.19. In Figure \ref{fig:seg}, we show the results on an example image from \cite{Funk_2017_ICCV_Workshops}. We observe that our method does not require any user interaction and still we get a comparable error rate. The performance of our method depends on how well the edges are extracted in the given image.
    			
    			\section{Conclusion}
    			\label{sec:c}
    			In this work, we have proposed an algorithm to partition an image into superpixels, such that the symmetry present at the pixel level is preserved at the superpixel level. We first detect the symmetry present at the pixel level, presented as pairs of mirror symmetric pixels and then extend the SLIC algorithm to preserve the symmetry at the superpixel level by proposing a novel symmetric initialization and symmetric pixel center assignment strategies. We observe that we are able to achieve mirror symmetric superpixels in the symmetric regions. We used detected pairs of mirror symmetric pixels to detect the symmetry axes present in the image. We also proposed an unsupervised symmetric objects segmentation approach using the SymmSLIC superpixels and achieve accuracy close to an unsupervised approach which requires human interface.  The main limitations of the proposed algorithm are that it is applicable only to the fronto-parallel views containing reflection symmetry and heavily depends on the performance of the edge detection algorithms. As a future work, we would like to extend the proposed method for the rotation symmetry, translation symmetry, and curved reflection symmetry. We also would like to prepare a dataset containing the set of ground truth pairs of superpixels for benchmarking the performance of the algorithm developed.
				\section*{Acknowledgement} Rajendra Nagar was supported by TCS Research Scholarship. Shanmuganathan Raman was supported by SERB-DST.
										

\end{document}